\pdfoutput=1
\PassOptionsToPackage{table}{xcolor}
\documentclass[10pt,logo,copyright]{nvidiatechreport}
\usepackage[authoryear,sort&compress,round]{natbib}
\usepackage{xurl}
\usepackage{parskip}
\usepackage{bm}
\usepackage{algorithm}
\usepackage{algpseudocode}
\usepackage{multirow}
\usepackage{wrapfig}
\usepackage{float}
\usepackage[section]{placeins}

\definecolor{modelrowcolor}{HTML}{D9F2B0}
\DeclareMathOperator*{\argmax}{arg\,max}
\title{VFlash: Fast, Accurate Visual Planning via Flash Primitive Decoding}
\author{
Yu-Cheng Chou$^{1*\dag}$,
Shihao Wang$^{2*\dag}$,
Jin Huang$^{*}$,
Chi-Pin Huang,
Chiao-An Yang$^{3\dag}$,
Xianzhe Fan$^{4\dag}$,
Jing Wang,
Zhiqi Li,
Yuanguo Kuang$^{2\dag}$,
Ruijie Zheng,
Johan Bjorck,
Sifei Liu,
Hang Su,
Abhishek Badki,
Lei~Zhang$^{2}$,
Alan Yuille$^{1}$,
Linxi ``Jim'' Fan,
Jan Kautz,
Zhiding Yu$^{\ddag}$
}

\newcommand{\authornotes}{%
\mbox{$^*$ Equal contribution. $^{\dag}$ Work done during an internship at NVIDIA. $^{\ddag}$ Corresponding author: \href{mailto:zhidingy@nvidia.com}{zhidingy@nvidia.com}.}\\
Additional affiliations: $^1$ Johns Hopkins University, $^2$ Hong Kong Polytechnic University,
$^3$ Purdue University, \mbox{$^4$ The University of Hong Kong.}
}

\hypersetup{pdftitle={VFlash: Fast, Accurate Visual Planning via Flash Primitive Decoding},pdfauthor={Yu-Cheng Chou, Shihao Wang, Jin Huang, Chi-Pin Huang, Chiao-An Yang, Xianzhe Fan, Jing Wang, Zhiqi Li, Yuanguo Kuang, Ruijie Zheng, Johan Bjorck, Sifei Liu, Hang Su, Abhishek Badki, Lei Zhang, Alan Yuille, Linxi Fan, Jan Kautz, Zhiding Yu}}

\begin{document}
\maketitle

\vspace{-8mm}

\noindent
\textbf{Links:}
{
\hypersetup{urlcolor=nvidiagreen}
\href{https://github.com/NVlabs/Eagle/tree/main/Embodied}{GitHub} |
\href{https://research.nvidia.com/labs/lpr/vflash/}{Project Page}
}

\begin{figure}[H]
    \centering
    \includegraphics[width=\linewidth]{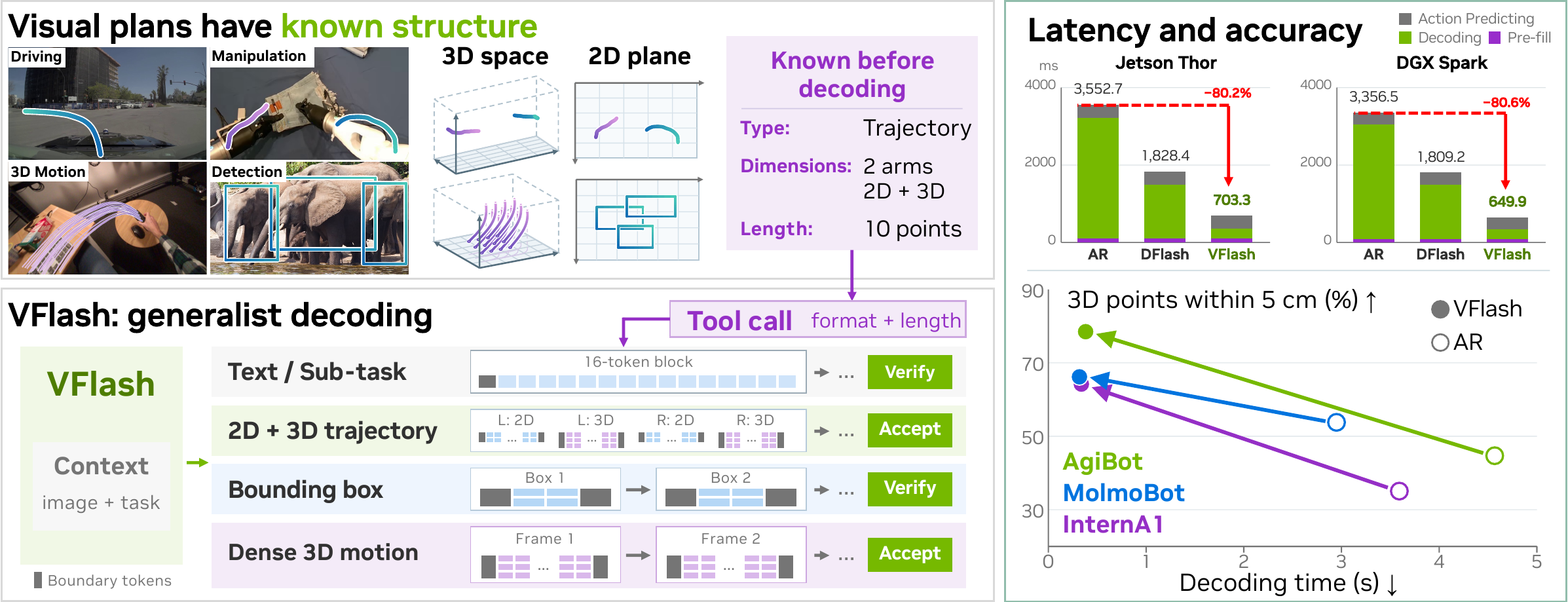}
    \captionsetup{font=small,skip=4pt}
    \caption{
    \textbf{VFlash} is a generalist decoding framework for fast, accurate visual planning across Physical AI tasks such as robotics and autonomous driving. By aligning prediction blocks with the known structure of visual primitives, it enables parallel decoding of common structured outputs such as bounding boxes, variable-length 2D/3D trajectories, and dense point motion, while retaining verified speculative decoding for text and sub-task plans. Compared with autoregressive (AR) decoding, VFlash improves accuracy on 3D robot trajectory planning while substantially reducing decoding latency across NVIDIA platforms, including Jetson Thor, DGX Spark, and H100.
    }
    \label{fig:teaser}
\end{figure}

\begin{abstract}
Models for physical AI tasks, such as robotics and autonomous driving, need not reason solely in language. They can use visual plans—points, bounding boxes, and trajectories—to represent spatial goals and anticipate motion. However, autoregressive vision-language models generate these plans one coordinate token at a time, making richer predictions increasingly costly. We introduce VFlash, a method for fast, accurate visual planning through Flash Primitive Decoding. VFlash exploits the known structure of visual primitives to predict their coordinates jointly, using a lightweight drafter that generates an entire primitive in one forward pass. The layout adapts to the primitive's type, dimensionality, and length, supporting compact boxes and long, variable-length trajectories. Crucially, spatial accuracy need not require reproducing an autoregressive model's exact token sequence. VFlash therefore trains on annotated primitives and directly accepts structurally valid predictions without target-token agreement. Across the evaluated 2D and 3D robot trajectory tasks, VFlash consistently reduces prediction errors relative to autoregressive decoding. Preliminary single-rollout measurements indicate approximately $25\times$ faster generation for 3D point motion forecasting. A world action model conditioned on VFlash plans achieves 70.2\% average success across four MolmoSpaces task suites, compared with 62.6\% for the baseline.
\end{abstract}

\abscontent

\clearpage
\setcounter{tocdepth}{2}
\addtocontents{toc}{\protect\setcounter{tocdepth}{-1}}

\section{Introduction}
\label{sec:introduction}

Vision-language models (VLMs) are increasingly used as high-level brains for physical AI systems, interpreting visual observations and task instructions to guide behavior in robotics and autonomous driving~\citep{team2025geminirobotics,drivevlm,nvidia2025alpamayo}. Their reasoning need not be expressed solely in language. VLMs can formulate intermediate predictions as visual plans: points and bounding boxes identify relevant locations or
objects~\citep{robopoint,rexomni}, while trajectories describe intended motion or forecast how a scene will evolve~\citep{rttrajectory,zhang2026molmomotion}. Recent work uses spatial traces and visual primitives to guide action generation~\citep{hamster,molmoact,guo2026avp}.
These plans make spatial goals and anticipated motion explicit before action.
Each visual primitive has identifiable boundaries and an organized
coordinate layout.

Generating such plans efficiently becomes increasingly important as their spatial detail grows.
Longer trajectories, additional objects, and joint 2D image and 3D
predictions require more coordinate tokens, which autoregressive VLMs
decode sequentially~\citep{chen2022pix2seq,bai2023qwenvlversatilevisionlanguagemodel,zhu2025internvl3,qwen3vl}.
For physical agents that update their plans as new observations arrive, sequential decoding limits spatial detail and update frequency under a fixed latency budget. This motivates decoding methods that can generate complete visual plans with fewer sequential forward passes while preserving their spatial accuracy.

Parallel decoding offers a promising route to reducing this sequential cost, particularly for latency-sensitive physical AI systems operating at batch size one. Existing approaches include speculative decoding with parallel 
verification~\citep{leviathan2023fast,li2025eagle3} and diffusion language models (dLLMs) with parallel denoising~\citep{nie2025llada,arriola2025block}. However, dLLMs often exhibit a speed–quality trade-off relative to autoregressive models~\citep{wu2025fast,liu2025sequential}. DFlash combines these approaches using a lightweight block diffusion draft model to predict draft tokens in a single forward pass, followed by target verification that preserves the autoregressive model’s output
distribution~\citep{chen2026dflash}. For visual outputs, LocateAnything exploits spatial structure through parallel box decoding~\citep{wang2026locateanything}.

In this work, we introduce \textbf{VFlash}, a generalist decoding framework for physical AI (Figure~\ref{fig:teaser}) that accommodates both text (e.g., sub-task plans) and diverse visual primitives, such as bounding boxes, variable-length trajectories, and dense 3D motion predictions. Our study reveals an interesting observation: a lightweight diffusion drafter can even surpass its autoregressive target in spatial accuracy of the visual plans while decoding much faster. The structure of visual primitives helps explain this finding: once a primitive’s type, dimensionality, and length are specified, its boundaries and coordinate layout are known. Aligning draft blocks with these boundaries allows an entire geometric object to be predicted jointly, with tens to hundreds of tokens decoded in a single forward pass. These observations motivate aligning training and decoding with complete primitive boundaries and directly accepting structurally valid drafts without requiring token-level agreement with the target, while retaining verified decoding for text.

At the center of VFlash is \textbf{Flash Primitive Decoding} (FPD), which uses the requested output structure to organize both drafter training and inference. A tool call specifies the size of a variable-length prediction, while the primitive type determines its coordinate layout. Conditioned on features from a frozen target VLM, a lightweight diffusion drafter learns to predict complete primitives from annotated spatial data. At inference, the decoder applies layout-specific constraints~\citep{willard2023outlines,geng2023grammar} and accepts valid primitives in full. The target VLM supplies context for subsequent predictions, and ordinary text retains speculative decoding with target verification.

We evaluate VFlash on 2D and 3D robot trajectories, bounding boxes, and dense 3D point motion forecasting, and study its use in downstream manipulation and driving. VFlash consistently reduces prediction errors relative to its autoregressive target across the evaluated robot trajectory tasks. In the downstream robot manipulation pipeline shown in Figure~\ref{fig:teaser}, it reduces inference latency by about 80\% on both Jetson Thor and DGX Spark. These visual plans also provide useful conditioning for action models: on MolmoSpaces~\citep{kim2026molmospaces}, a world action model~\citep{ye2026dreamzero,nvidia2026worldsimulation} conditioned on VFlash plans achieves 70.2\% average success across four task suites, compared with 62.6\% for the baseline.

Beyond acceleration, our work suggests a broader role for parallel decoding in multimodal systems. Lossless speculative decoding preserves the autoregressive target’s output distribution~\citep{leviathan2023fast,chen2023accelerating}, whereas structured visual planning instead permits predictions to be judged by their spatial accuracy and usefulness for action, in the spirit of relaxed acceptance criteria~\citep{bachmann2025judge,wang2025specvla}. VFlash enables a System-1 mode~\citep{kahneman2011thinking} of visual planning in VLMs, producing fast, structured spatial predictions through joint coordinate decoding. This perspective interestingly connects to recent System-1 models such as Jev, which emphasize fast, typed decisions~\citep{almeida2026systemone,yu2026visualjev}. VFlash brings this perspective to complete geometric predictions, exploiting structure within each primitive. The resulting gains in both speed and spatial accuracy suggest that autoregressive decoding need not define the quality ceiling for visual outputs.

\section{Method}
\label{sec:method}

\begin{figure}[!t]
    \centering
    \includegraphics[width=\linewidth]{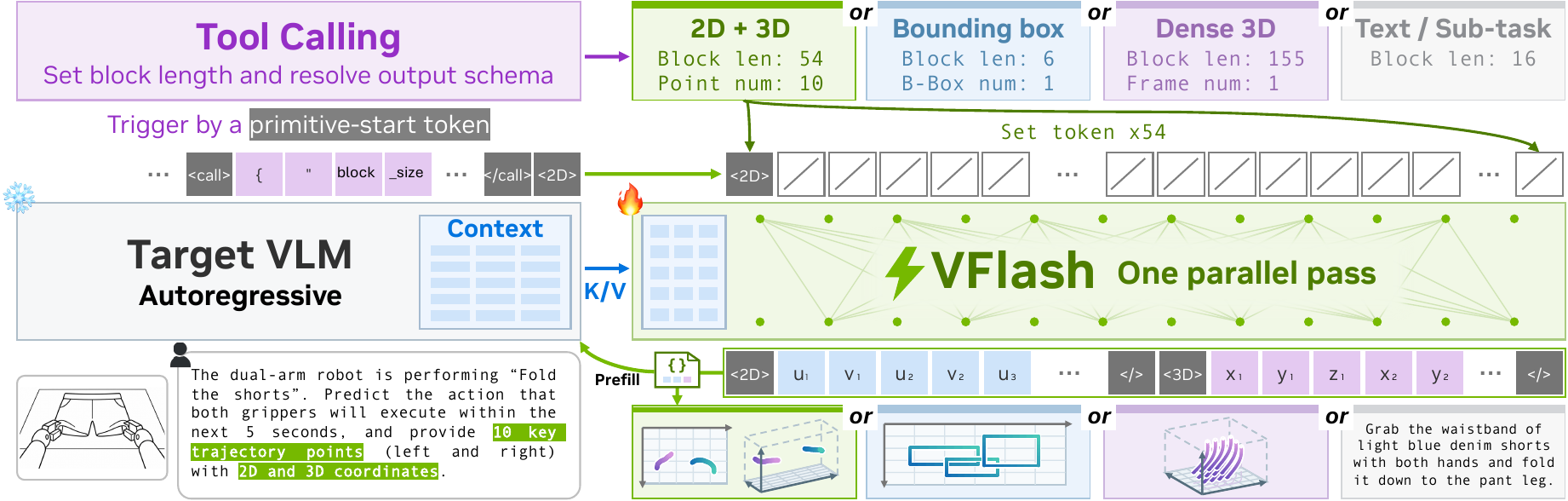}
    \captionsetup{font=small,skip=4pt}
    \caption{\textbf{FPD design.}
    A primitive-start token triggers a prediction block layout determined by the primitive type and the preceding tool calling. By adapting the prediction block to each primitive's structure, VFlash supports different spatial formats and lengths without requiring token-level agreement with the VLM.
    }
    \label{fig:pipeline}
\end{figure}

This section presents the generalist decoding framework of VFlash for physical AI, 
covering both the preliminaries of DFlash and the detailed design of Flash
Primitive Decoding (Figure~\ref{fig:pipeline}).

\subsection{Preliminaries}
\label{sec:preliminaries}

In speculative decoding, a draft model $q_\phi$ proposes $\gamma$ tokens
that the target model $p_\theta$ verifies in parallel, giving an average
per-token latency of
\begin{equation}
    L=\frac{T_{\mathrm{draft}}+T_{\mathrm{verify}}}{\tau},
    \label{eq:spec_latency}
\end{equation}
where $\tau$ is the expected number of tokens accepted per cycle~\citep{chen2026dflash}. DFlash reduces the numerator with a lightweight block diffusion drafter. Features $H_\theta(x)$ from selected target layers are fused and injected as keys and values into each draft layer. Given a start token $a_0$, the drafter predicts $\gamma$ masked tokens in one bidirectional forward pass, yielding $T_{\mathrm{draft}}=t_{\mathrm{par}}\ll\gamma\,t_{\mathrm{step}}$, where $t_{\mathrm{step}}$ is the sequential drafting cost per token. Longer drafts, however, need not increase $\tau$: greedy verification accepts tokens only up to the first mismatch with the target:
\begin{equation}
    \tau_{\mathrm{ver}}=1+\max\Big\{k\le\gamma:\;
    a_j=\argmax_{v}\,p_\theta\!\left(v\mid x,a_{<j}\right)
    \ \text{for all}\ j\le k\Big\}.
    \label{eq:verified_acceptance}
\end{equation}
$\tau_{\mathrm{ver}}$ often saturates well below $\gamma$. Thus enlarging a block does not necessarily
enlarge its final throughput. For text generation, the goal is to preserve the target distribution~\citep{leviathan2023fast} via verification. But token-level agreement need not reflect spatial accuracy. VFlash retains the drafting architecture and replaces Eq.~\ref{eq:verified_acceptance} with direct acceptance of structurally valid visual primitives.

\subsection{Flash primitive decoding}
\label{sec:problem_setup}

Given visual observations and a task instruction, the VLM predicts spatial objects such as boxes, robot trajectories, or future point positions. Each prediction is a \textit{visual primitive} with a schema $s$ specifying its type, coordinate dimensions, and extent. This fixes each coordinate’s role while leaving its value to be inferred from the input. A trajectory’s 2D image coordinates and 3D spatial coordinates can form one primitive and be predicted jointly. For variable-length predictions, a tool call declares the extent, which together with the primitive type determines the layout; for fixed-size predictions such as boxes, the type alone suffices. The structure is thus known before coordinate prediction, and each decoding block covers a complete primitive.

The schema defines a block of length $B(s)$: an observed start token
$a_0$ followed by $B(s)-1$ masked tokens. An autoregressive target
factorizes the blocks as $\prod_j p_\theta(a_j\mid x,a_{<j})$ and needs
$B(s)-1$ sequential passes. Given the layout and the target context,
VFlash instead predicts all positions conditionally independently,
\begin{equation}
    q_\phi\!\left(a_{1:B(s)-1}\mid H_\theta(x),a_0\right)
    =\prod_{j=1}^{B(s)-1}
     q_{\phi,j}\!\left(a_j\mid H_\theta(x),a_0,
     \mathrm{MASK}^{B(s)-1}\right),
    \label{eq:primitive_factorization}
\end{equation}
in a single forward pass. The complete primitive is thus the unit of
parallel prediction, with a speculation budget $\gamma=B(s)-1$ set by the
request rather than by a fixed window. The same principle covers targets
that emit coordinates as ordinary text: we align their numeric fields in
a fixed-width layout and render the predictions back into the target's
text format. Appendix~\ref{app:schemas} gives the coordinate conventions
and serialization formats of each task.

\subsection{Token representations}
\label{sec:token_representations}

Visual plans often require large, variable-length draft blocks.
For example, NAVSIM requires eight waypoints over four seconds,
while Waymo requires twenty over five seconds. Drafting such blocks
introduces challenges in format following and spatial accuracy.
We find that two design principles help address these challenges:

\textbf{Boundary alignment} decides what is predicted together:
each draft block covers one complete primitive, avoiding the splits
caused by fixed-length windows. The tool call declares the block
length, from which the waypoint count $P$ follows, and a shared layout
template accommodates different lengths, allowing the whole primitive
to be drafted in one pass.

\textbf{Position alignment} decides what each position predicts:
every slot has a known role---a waypoint, axis, or delimiter---before
decoding. This supports parallel prediction with a consistent layout
and per-position constraints at inference.

\begin{table}[H]
\centering
\captionsetup{font=small,skip=4pt}
\caption{\textbf{Primitive layouts.} 
Each block begins with the start token $a_0$. $P$ denotes the number of points;
$(u,v)$ and $(x,y,z)$ denote image-plane and metric coordinates in special tokens.
}
\footnotesize
\setlength{\tabcolsep}{8pt}
\renewcommand{\arraystretch}{1.15}
\begin{tabular}{ll}
\toprule
\textbf{Primitive} & \textbf{Layout} \\ %
\midrule
B-box
  & \texttt{<box>}\,$u_1\,v_1\,u_2\,v_2$\,\texttt{</>}
  \\%
2D+3D traj.
  & \texttt{<2D>}\,$(u\,v)\!\!\times\!\!{P}$\,\texttt{</>}\;\texttt{<3D>}\,$(x\,y\,z)\!\!\times\!\!{P}$\,\texttt{</>}
  \\%
Dense motion
   & \texttt{<Dense>}\;$t_1 t_2$\;$\big[\,k\;(s\,d_1{\cdots}d_5)\!\!\times\!\!{3}\,\big]\!\!\times\!\!{P}$\texttt{</>} \\
\bottomrule
\end{tabular}
\label{tab:primitive_formats}
\end{table}

Maintaining this layout requires each coordinate to occupy a fixed
number of positions. Recent VLMs normalize image
coordinates to 0--1,000 and write them as ordinary
text~\citep{bai2023qwenvlversatilevisionlanguagemodel,zhu2025internvl3};
models built around visual primitives often use special
tokens~\citep{rexomni,wang2026locateanything}. A special coordinate
token occupies one position, whereas text tokenization can vary
with the value and shift subsequent coordinates. VFlash supports
both representations.

For our base model (Section~\ref{sec:base_model}), typed start and
end tokens mark primitive boundaries and transitions between text
and primitive decoding. Each coordinate uses one of 1,001 special
tokens, allowing prediction to be restricted to its legal vocabulary. 
For text-tokenized targets, VFlash gives each coordinate a sign and
five digit positions, with unused high-order positions filled by the
mask token. This fixed-width grid preserves position roles and is
rendered back into the target's text format, whose delimiters mark
block boundaries. Table~\ref{tab:primitive_formats} and Appendix~\ref{app:schemas} give the formats details.

\subsection{Training}
\label{sec:primitive_training}

We adapt an autoregressive VLM to the spatial response format with next-token supervision, then freeze it and train a shallow, non-causal drafter on its features. For an annotated block $a$, we minimize the weighted cross-entropy of Eq.~\ref{eq:primitive_factorization},
\begin{equation}
    \mathcal{L}_{\mathrm{prim}}
    = -\sum_{j=1}^{B(s)-1} w_j
      \log q_{\phi,j}\!\left(a_j\mid
      H_\theta(x),a_0,\mathrm{MASK}^{B(s)-1}\right),
    \label{eq:primitive_training}
\end{equation}
with all logits computed jointly in one forward pass.
Primitive positions use uniform $w_j$, unlike the decaying weights
that favor early positions under Eq.~\ref{eq:verified_acceptance}.
Crucially, labels $a_j$ come from annotated coordinates or rendered
grids rather than the target's continuation, optimizing the drafter
for spatial accuracy instead of token-level agreement.

Each annotated primitive contributes one complete block. Where the response also contains text, sampled short blocks train the ordinary drafting path, so one drafter serves both the verified text path and the direct primitive path. Packed attention keeps examples and independent blocks separate while preserving attention within each primitive. Backbones, data, validity weights, and adaptation recipes are given in Appendix~\ref{app:upstream_training}.

\subsection{Inference}
\label{sec:flash_primitive_decoding}

Decoding runs 16-position DFlash blocks until a primitive-start token is
accepted; the decoder then resolves the declared schema and drafts the
complete block of length $B(s)$ in one pass. Per-position constraints
restrict coordinates and delimiters to their legal vocabularies, and
cross-position rules pair out-of-frame coordinates; both act on logits
already computed and add no neural evaluation.

Let $\hat{a}$ be the resulting prediction and $A_s$ its acceptance test:
the schema validator, plus for boxes the spatial-confidence guard of
LocateAnything~\citep{wang2026locateanything}. VFlash replaces
Eq.~\ref{eq:verified_acceptance} with
\begin{equation}
    \tau_{\mathrm{dir}}=
    \begin{cases}
        B(s), & A_s(\hat{a})=1,\\[2pt]
        1+r, & \text{otherwise},
    \end{cases}
    \label{eq:direct_acceptance}
\end{equation}
where the first $r$ draft slots are retained ($r=0$ for trajectories;
for boxes, those before the first uncertain coordinate) and the rest are
regenerated by constrained target AR. An accepted primitive is committed
without token-level agreement; one target pass updates the cache and
context features and emits the next token. Text still follows
Eq.~\ref{eq:verified_acceptance}; Appendix~\ref{app:decoding_algorithm}
gives both procedures.

Substituting Eq.~\ref{eq:direct_acceptance} into
Eq.~\ref{eq:spec_latency}, a valid primitive costs
\begin{equation}
    L_{\mathrm{dir}}
    =\frac{t_{\mathrm{par}}+t_{\mathrm{target}}}{B(s)},
    \label{eq:primitive_latency}
\end{equation}
against one target forward $t_{\mathrm{target}}$ per coordinate under
autoregressive decoding, up to prefix and continuation overhead. Both
passes grow mildly with $B(s)$, but their number does not, so a more
detailed plan adds no sequential decoding steps.

\section{Experiment}
\label{sec:experiments}

\textbf{Base VLM.}\phantomsection\label{sec:base_model}
Most experiments use a customized embodied VLM combining
Qwen3-4B~\citep{qwen3}, a native-resolution MoonViT
encoder~\citep{kimivl}, and an MLP connector. It is trained
progressively on general vision-language tasks,
grounding~\citep{wang2026locateanything}, embodied spatial
understanding, and joint 2D/3D trajectory prediction.
Coordinates for bounding box and trajectory use special tokens, with 2D image coordinates normalized
to 0--1,000 and 3D coordinates uniformly quantized within
predefined ranges. For dense motion prediction, we use the
released MolmoMotion-AR model~\citep{zhang2026molmomotion},
retaining its text-based coordinate representation.

\textbf{Per-task training.}
For clean ablation studies, we fine-tune the AR target on
task-specific data, then freeze it and train a five-layer VFlash
drafter on the same data. For MolmoMotion, we train only the
drafter on the frozen released target. Detailed per-task datasets,
training mixtures, and optimization settings are provided in
Appendix~\ref{app:upstream_training}.

\textbf{Visual-plan conditioning.}
Despite differences in DiT/WAM architectures and their interfaces
with the VLM, a commonality is that the action models cross-attend
to VLM features encoding the generated visual plans.
These features are obtained by processing VFlash's accepted
primitives with the frozen target VLM.
Appendix~\ref{app:action_models} details the architectures and
conditioning mechanisms.

\begin{table*}[t]
\centering
\begin{minipage}[t]{0.54\textwidth}
\vspace{0pt}
\centering
\captionsetup{font=scriptsize,labelfont=bf,justification=raggedright,singlelinecheck=false,skip=3pt}
\caption{\textbf{Open-loop 2D trajectory prediction.}
DFD, HD, and RMSE in normalized image coordinates ($\downarrow$).
Reference rows are taken from prior work; Baseline (AR) and VFlash
share the same 4B target and differ only in the decoder. Bold and
underline mark the best and second-best values.}
\label{tab:open_loop_2d}
\scriptsize
\renewcommand{\arraystretch}{1.21}
\setlength{\tabcolsep}{2.5pt}
\resizebox{\linewidth}{!}{%
\begin{tabular}{lc*{6}{c}}
\toprule
& & \multicolumn{3}{c}{\textbf{ShareRobot-Bench}} & \multicolumn{3}{c}{\textbf{VABench-V}} \\
\cmidrule(lr){3-5}\cmidrule(lr){6-8}
\textbf{Method} & \textbf{Params}
& DFD & HD & RMSE
& DFD & HD & RMSE \\
\midrule
Qwen3-VL~\citeyearpar{qwen3vl} & 4B & 0.381 & 0.329 & 0.220 & 0.279 & 0.253 & 0.204 \\
Qwen3-VL~\citeyearpar{qwen3vl} & 8B & 0.392 & 0.341 & 0.233 & 0.274 & 0.255 & 0.202 \\
\midrule
MolmoAct~\citeyearpar{molmoact} & 7B & 0.776 & 0.776 & 0.677 & 0.814 & 0.814 & 0.688 \\
Embodied-R1~\citeyearpar{embodiedr1} & 3B & 0.343 & 0.300 & 0.239 & 0.303 & 0.259 & 0.213 \\
HAMSTER~\citeyearpar{hamster} & 13B & 0.437 & 0.392 & 0.355 & 0.212 & 0.205 & 0.183 \\
RoboBrain-2.0~\citeyearpar{baairobobrainteam2025robobrain20technicalreport} & 7B & \textbf{0.167} & \textbf{0.158} & \textbf{0.125} & 0.329 & 0.260 & 0.224 \\
LoHo-Manip~\citeyearpar{lohomanip} & 4B & 0.231 & 0.206 & 0.156 & 0.212 & 0.182 & 0.147 \\
\midrule
Baseline (AR) & 4B & 0.202 & 0.175 & 0.142
& \underline{0.130} & \underline{0.125} & \underline{0.093} \\
\rowcolor{modelrowcolor}
\textbf{VFlash (ours)} & 4B & \underline{0.194} & \underline{0.162} & \underline{0.133}
& \textbf{0.126} & \textbf{0.116} & \textbf{0.089} \\
\bottomrule
\end{tabular}%
}

\end{minipage}\hfill
\begin{minipage}[t]{0.43\textwidth}
\vspace{0pt}
\centering
\captionsetup{font=scriptsize,labelfont=bf,justification=raggedright,singlelinecheck=false,skip=3pt}
\caption{\textbf{Category-conditioned box generation.}
F1 (\%, $\uparrow$) and boxes per second (BPS, $\uparrow$).
\emph{Direct} accepts every schema-valid box without verification;
\emph{AR fallback} regenerates boxes rejected by the spatial-confidence
guard with constrained AR (Appendix~\ref{app:box_protocols}).
BPS is measured separately on each dataset.}
\label{tab:box_generation}
\vfill

\scriptsize
\renewcommand{\arraystretch}{1.3}
\setlength{\tabcolsep}{2.5pt}

\begin{tabularx}{\linewidth}{ll*{4}{>{\centering\arraybackslash}X}}
\toprule
& & \multicolumn{3}{c}{\textbf{F1@IoU (\%)}} & \\
\cmidrule(lr){3-5}
\textbf{Method} & \textbf{Dataset} & Mean & 0.50 & 0.95 & BPS \\
\midrule
& COCO
    & \textbf{48.68} & 65.29 & \textbf{16.09} & 3.15 \\
\multirow{-2}{*}{Baseline (AR)} & Dense200
    & 60.10 & 76.70 & \textbf{16.95} & 6.73 \\
\midrule
& COCO
    & 41.32 & 59.64 & 10.76 & 5.96 \\
\multirow{-2}{*}{\shortstack[l]{VFlash\\(direct)}} & Dense200
    & 39.09 & 55.06 & 6.59 & \textbf{22.66} \\
\midrule
\rowcolor{modelrowcolor}
& COCO
    & 47.79 & \textbf{66.38} & 13.48 & 8.76\\
\rowcolor{modelrowcolor}
\multirow{-2}{*}{\shortstack[l]{\textbf{VFlash}\\\textbf{(AR fallback)}}} & Dense200
    & \textbf{61.78} & \textbf{82.64} & 12.11 & 17.46 \\
\bottomrule
\end{tabularx}

\end{minipage}
\end{table*}

\begin{table}[t]
\centering
\captionsetup{font=small,skip=4pt}
\caption{\textbf{Open-loop 3D robot trajectory prediction.}
ADE and FDE in meters ($\downarrow$); SR@5 and SR@10 are the fraction of
waypoints within 5\,cm and 10\,cm ($\uparrow$). The three variants differ in what the
model generate (3D only, 2D+3D) and whether wrist-camera views were
added in training. All groups are evaluated with the same single-view
inputs. Bold marks the best value per column.}
\label{tab:robot_3d_trajectories}
\small
\renewcommand{\arraystretch}{1.13}
\setlength{\tabcolsep}{8pt}
\resizebox{1\linewidth}{!}{%
\begin{tabular}{ll*{12}{c}}
\toprule
& & \multicolumn{4}{c}{\textbf{InternA1}~\citeyearpar{tian2026interndata}}
& \multicolumn{4}{c}{\textbf{AgiBot}~\citeyearpar{agibot}}
& \multicolumn{4}{c}{\textbf{MolmoBot}~\citeyearpar{deshpande2026molmobot}} \\
\cmidrule(lr){3-6}\cmidrule(lr){7-10}\cmidrule(lr){11-14}
\textbf{Training Variant} & \textbf{Decoder}
& ADE & FDE & SR@5 & SR@10
& ADE & FDE & SR@5 & SR@10
& ADE & FDE & SR@5 & SR@10 \\

\midrule
& AR
& 0.101 & 0.155 & 0.412 & 0.658
& 0.083 & 0.125 & 0.503 & 0.799
& 0.066 & 0.104 & 0.526 & 0.804 \\
\rowcolor{modelrowcolor}
\cellcolor{white}\multirow{-2}{*}{\textbf{3D-only}}
& \textbf{VFlash}
& 0.065 & 0.098 & 0.638 & 0.820
& 0.042 & 0.067 & 0.731 & 0.914
& 0.054 & 0.083 & 0.653 & 0.870 \\

\midrule
& AR
& 0.153 & 0.233 & 0.273 & 0.473
& 0.105 & 0.151 & 0.398 & 0.698
& 0.076 & 0.116 & 0.467 & 0.767 \\
\rowcolor{modelrowcolor}
\cellcolor{white}\multirow{-2}{*}{\textbf{2D+3D}}
& \textbf{VFlash}
& 0.062 & 0.091 & 0.640 & 0.827
& 0.041 & 0.064 & 0.739 & 0.919
& 0.055 & 0.085 & 0.640 & 0.862 \\

\midrule
& AR
& 0.130 & 0.196 & 0.360 & 0.554
& 0.084 & 0.129 & 0.456 & 0.749
& 0.066 & 0.102 & 0.547 & 0.813 \\
\rowcolor{modelrowcolor}
\cellcolor{white}\multirow{-2}{*}{%
  \shortstack[l]{\textbf{2D+3D w/ wrist-}\\\textbf{camera training}}}
& \textbf{VFlash}
& \textbf{0.060} & \textbf{0.091} & \textbf{0.651} & \textbf{0.835}
& \textbf{0.036} & \textbf{0.057} & \textbf{0.794} & \textbf{0.937}
& \textbf{0.052} & \textbf{0.080} & \textbf{0.671} & \textbf{0.877} \\
\bottomrule
\end{tabular}%
}
\end{table}

\subsection{Visual primitive prediction}
\label{sec:upstream}

\textbf{2D trajectory prediction.}
On ShareRobot-Bench~\citep{ji2025robobrain} and
VABench-V~\citep{yuan2026seeing}, VFlash lowers all six errors
of its AR target (Table~\ref{tab:open_loop_2d}), by 7\% on HD
and 3--6\% on DFD and RMSE across both benchmarks.
On VABench-V, VFlash cuts every error by roughly 40\% relative
to the best prior model, LoHo-Manip~\citep{lohomanip},
with our AR target already outperforming the listed baselines;
on ShareRobot-Bench, VFlash is second only to
RoboBrain-2.0~\citep{baairobobrainteam2025robobrain20technicalreport}.

\textbf{2D bounding box prediction.}
On COCO~\citep{lin2014microsoft} and Dense200~\citep{rexomni}, VFlash
with AR fallback matches AR on mean F1 while decoding
2.6--2.8$\times$ faster (Table~\ref{tab:box_generation}), with higher
F1 at IoU 0.50, especially on the crowded Dense200 scenes, but lower F1
at IoU 0.95. Without fallback, mean F1 drops by about 20 points on
Dense200.

\textbf{3D trajectory prediction.}
On InternA1~\citep{tian2026interndata}, AgiBot~\citep{agibot}, and
MolmoBot~\citep{deshpande2026molmobot}, VFlash improves all four
metrics for all three target variants
(Table~\ref{tab:robot_3d_trajectories}). ADE falls by 18--49\% with the
3D-only target and by 28--61\% with the 2D+3D target; waypoint
accuracy at 5\,cm on AgiBot rises from 50\% to 73\% and from
40\% to 74\%, respectively.
With joint 2D+3D outputs, AR's 3D ADE is 15--51\% higher than
with 3D-only outputs, while VFlash maintains similar accuracy
by predicting both segments in one parallel pass.
The wrist-camera training variant further reduces VFlash's ADE
by 3--12\%, with evaluation still using only the primary camera.

\textbf{Dense 3D motion prediction.}
On PointMotionBench~\citep{zhang2026molmomotion}, we freeze the backbone of
MolmoMotion-AR and draft each future frame of query points as one
block with a fixed-width digit layout (Appendix~\ref{app:schemas}).
As shown in Table~\ref{tab:molmomotion}, VFlash reduces DAVIS ADE and FDE by 27\%
and 19\%, respectively, and HOT3D FDE by 8\%.
Its ADE ranks second on HOT3D and WorldTrack, behind AR but ahead
of the flow-matching variant, although WorldTrack FDE increases
by 15\% and PWT is 7--12\% lower across the three datasets.
These results extend VFlash to dense motion prediction with
text-tokenized coordinates, with gains over AR on several error
metrics and approximately $25\times$ faster single-rollout generation
(Appendix~\ref{app:molmomotion_protocol}).

\begin{table}[t]
\centering
\captionsetup{font=small,skip=4pt}
\caption{
\textbf{3D point motion forecasting on PointMotionBench.}
VFlash uses MolmoMotion-AR~\citep{zhang2026molmomotion} as its target.
VFlash achieves approximately $25\times$ faster single-rollout
generation than AR (Appendix~\ref{app:molmomotion_protocol}).
ADE/FDE are in meters ($\downarrow$); Frames and Text indicate observation history and language input.
Bold and underline mark the best and second-best values.
}
\label{tab:molmomotion}
\small
\renewcommand{\arraystretch}{1.08}
\setlength{\tabcolsep}{8pt}
\definecolor{molmomotionrefrowcolor}{HTML}{FFF2CC}
\resizebox{\linewidth}{!}{%
\begin{tabular}{lcc*{9}{c}}
\toprule
& \multicolumn{2}{c}{\textbf{Input}}
& \multicolumn{3}{c}{\textbf{HOT3D}}
& \multicolumn{3}{c}{\textbf{WorldTrack}}
& \multicolumn{3}{c}{\textbf{DAVIS}} \\
\cmidrule(lr){2-3}\cmidrule(lr){4-6}\cmidrule(lr){7-9}\cmidrule(lr){10-12}
\textbf{Method} & \textbf{Frames} & \textbf{Text}
& ADE$\downarrow$ & FDE$\downarrow$ & PWT$\uparrow$
& ADE$\downarrow$ & FDE$\downarrow$ & PWT$\uparrow$
& ADE$\downarrow$ & FDE$\downarrow$ & PWT$\uparrow$ \\
\midrule
ObjectForesight~\citeyearpar{soraki2026objectforesight} & 3 & No & 0.129 & \underline{0.192} & 0.353 & --- & --- & --- & --- & --- & --- \\
EgoScaler~\citeyearpar{yoshida2025generating} & 1 & Yes & 0.170 & \textbf{0.179} & 0.218 & --- & --- & --- & --- & --- & --- \\
Robot4DGen~\citeyearpar{liu2026geometry} & 3 & No & 0.212 & 0.271 & 0.112 & 0.548 & 0.704 & 0.121 & 2.120 & 3.382 & 0.081 \\
\midrule
Track2Act~\citeyearpar{bharadhwaj2024track2act} & 1 & No & 0.294 & 0.413 & 0.202 & 1.230 & 1.567 & 0.053 & 4.853 & 8.110 & 0.018 \\
\midrule
MolmoMotion-FM~\citeyearpar{zhang2026molmomotion} & 3 & Yes & 0.135 & 0.255 & 0.382 & 0.158 & \underline{0.295} & \underline{0.438} & 1.480 & 2.520 & 0.130 \\
MolmoMotion-AR~\citeyearpar{zhang2026molmomotion} & 3 & Yes & \textbf{0.109} & 0.217 & \textbf{0.444} & \textbf{0.143} & \textbf{0.261} & \textbf{0.445} & 1.227 & 2.108 & \textbf{0.153} \\
\midrule
\rowcolor{modelrowcolor}
\textbf{VFlash} & 3 & Yes & \underline{0.122} & 0.200 & \underline{0.391} & \underline{0.145} & 0.300 & 0.408 & \textbf{0.893} & \textbf{1.716} & \underline{0.143} \\
\bottomrule
\end{tabular}%
}
\end{table}

\subsection{Downstream applications}
\label{sec:applications}
\label{sec:plan_conditioning}

\textbf{Experiment Setup.} 
We prioritize controlled comparisons of visual-plan conditioning over maximizing benchmark scores through additional model enhancements. The VLM remains frozen in all configurations; action policies condition on its hidden states encoding the generated plans and visual context. Within each application, we keep the action-model
architecture fixed across comparisons.

\begin{table}[t]
\centering

\begin{minipage}[t]{0.545\textwidth}
\vspace{0pt}
\centering
\captionsetup{font=small,skip=4pt}
\caption{\textbf{MolmoSpaces manipulation results.}
Baseline and VFlash share the same WAM; Baseline keeps its original
text encoder, VFlash replaces it with our VLM and conditions on the
generated plan. Bold marks the best value per column.}\label{tab:molmospaces_main}

\scriptsize
\renewcommand{\arraystretch}{1.1}
\setlength{\tabcolsep}{4.5pt}
\resizebox{\linewidth}{!}{%
\begin{tabular}{lccccc}
\toprule
\textbf{Method}
    & \textbf{Avg.}
    & \textbf{Pick}
    & \textbf{P\&P}
    & \textbf{Open}
    & \textbf{Close} \\
\midrule

$\pi_0$-FAST DROID~\citeyearpar{pertsch2025fastefficientactiontokenization}
    & 20.7 & 22.0 & 11.0 & 11.1 & 38.6 \\

$\pi_0$ DROID~\citeyearpar{pi0}
    & 23.2 & 16.2 & 12.5 & 11.0 & 53.1 \\

$\pi_{0.5}$ DROID~\citeyearpar{intelligence2025pi05visionlanguageactionmodelopenworld}
    & 34.5 & 36.4 & 13.6 & 22.7 & 65.1 \\

PRTS-Droid~\citeyearpar{zhang2026prtsprimitivereasoningtasking}
    & 42.4 & 53.8 & 27.7 & 12.0 & 76.3 \\

Cosmos3-Nano-Policy~\citeyearpar{nvidia2026cosmos3omnimodalworld}
    & 53.0 & 65.5 & 35.2 & 32.0 & 79.1 \\

WALL-OSS-0.5~\citeyearpar{yu2026walloss05technicalreport}
    & 53.8 & 73.8 & 46.0 & 16.8 & 78.5 \\

PrimeR0~\citeyearpar{lumos2026primer0}
    & 60.3 & 77.5 & 54.6 & 27.8 & \textbf{81.3} \\

\midrule
Baseline
    & 62.6 & 81.1 & 55.7 & 49.3 & 64.2 \\

\rowcolor{modelrowcolor}
\textbf{VFlash}
    & \textbf{70.2}
    & \textbf{83.6}
    & \textbf{67.3}
    & \textbf{52.8}
    & 77.0 \\

\bottomrule
\end{tabular}%
}
\end{minipage}%
\hfill
\begin{minipage}[t]{0.425\textwidth}
\vspace{0pt}
\centering
\captionsetup{font=small,skip=4pt}
\caption{
All variants use the same WAM with different
decoders, primitive types, or conditioning features.
Baseline uses WAM's original text encoder.
Q, S, T, and I denote the VLM hidden features for the question,
subtask, trajectory, and image tokens.
Image context baseline uses VFlash to decode subtask.
}
\label{tab:molmospaces_ablation}
\scriptsize
\renewcommand{\arraystretch}{1.1}
\setlength{\tabcolsep}{2.5pt}
\resizebox{\linewidth}{!}{%
\begin{tabular}{lcccc}
\toprule
\textbf{Variant}
    & \textbf{Decode}
    & \textbf{Primitive}
    & \textbf{Context}
    & \textbf{Avg.} \\
\midrule

Baseline
    & --- & --- & Native
    & 62.6 \\

AR generation
    & AR & 2D+3D & Q+S+T
    & 61.0 \\

3D-only
    & VFlash & 3D & Q+S+T
    & 69.3 \\

Image context
    & VFlash & --- & Q+S+I
    & 63.7 \\

\midrule
\rowcolor{modelrowcolor}
\textbf{VFlash}
    & VFlash & 2D+3D & Q+S+T
    & \textbf{70.2} \\

\bottomrule
\end{tabular}%
}
\end{minipage}
\end{table}

\textbf{MolmoSpaces.}
For MolmoSpaces~\citep{kim2026molmospaces} the policy is a world
action model (WAM) obtained by post-training
Cosmos-Predict2.5~\citep{nvidia2026worldsimulation} for robot control.
We replace its Cosmos-Reason text encoder with our VLM. The
WAM reads the VLM's hidden states at the question, the generated
subtask, and the visual plans of 2D+3D trajectories with ten waypoints, passed through a linear
projection into its existing text cross-attention path. 
It keeps its own camera and robot-state inputs, 
predicts 20-step action chunks, and requests a new
plan every three chunks. The baseline is the unmodified WAM with its
original text encoder.

As shown in Table~\ref{tab:molmospaces_main}, visual-plan conditioning
raises average success from 62.6\% to 70.2\%, exceeding the
listed public model, PrimeR0~\citep{lumos2026primer0}, by 9.9 percentage
points on average and scoring higher on three of four tasks.
The largest gains over the baseline are on Pick \& Place and Close
(11.6 and 12.8 points), compared with 2.5 points on Pick,
suggesting greater benefits for tasks involving multiple dependent motions.

Table~\ref{tab:molmospaces_ablation} compares plan representations
and conditioning choices. Using AR-generated 2D+3D plans instead
of VFlash plans lowers average success from 70.2\% to 61.0\%.
Replacing trajectory states with image states lowers success
by 6.5 percentage points to 63.7\%.
A 3D-only plan reaches 69.3\%, while joint 2D+3D plans achieve
the best average.

\begin{figure}[!htbp]
    \centering
    \includegraphics[width=0.65\linewidth]{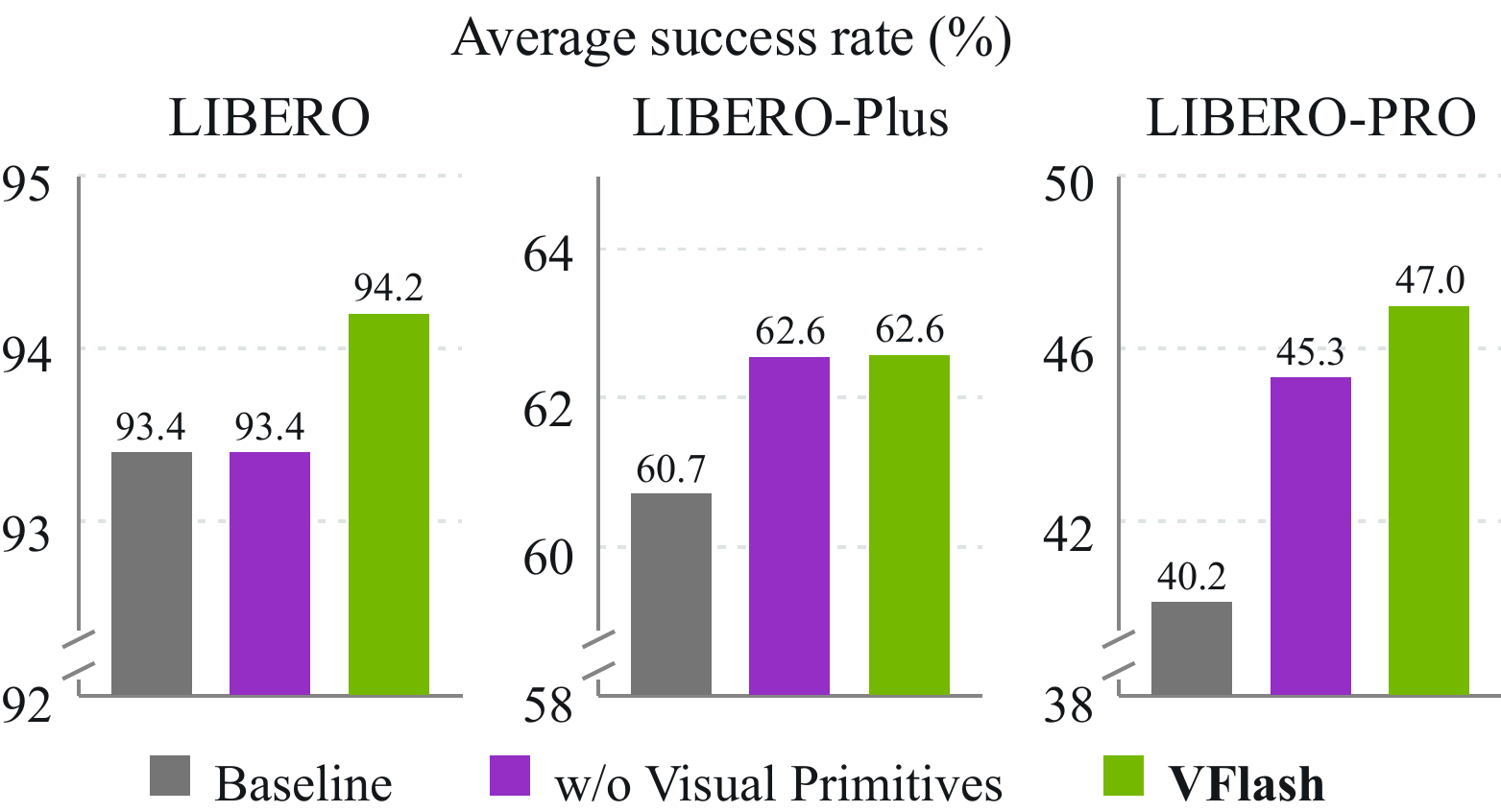}
    \captionsetup{font=small,skip=4pt}
    \caption{\textbf{LIBERO family.} Average success (\%) of the same
    PIv3 head conditioned on a frozen Qwen3-VL-4B (Baseline), on our
    VLM's states up to the generated subtask (w/o Visual Primitives),
    and on the full visual plan (VFlash).}
    \label{fig:exp_wrap}
\end{figure}

\textbf{LIBERO family.}
On LIBERO~\citep{libero}, LIBERO-Plus~\citep{liberoplus}, and
LIBERO-PRO~\citep{zhou2025liberopro}, we train a PIv3 flow-matching
action head in StarVLA~\citep{community2026starvla}, with per-layer
cross-attention to a frozen VLM.
Given primary and wrist images, the VLM generates a subtask and
a ten-waypoint 2D+3D trajectory before each eight-action chunk.
We compare conditioning on VLM states through the subtask or
the full plan. The baseline uses the same action head with a frozen
Qwen3-VL-4B encoding only images and instructions.

As shown in Figure~\ref{fig:exp_wrap}, gains over the baseline increase
from 0.8 percentage points on LIBERO to 1.9 on LIBERO-Plus and 6.8
on LIBERO-PRO, indicating greater benefits under perturbations.
On LIBERO-PRO, visual plans improve Spatial, Object, and Goal by
11.1, 9.4, and 6.9 points over the baseline.
Compared with subtask-only conditioning, they add 2.6 points on
Spatial, 1.2 on Goal, and 2.0 on Long, showing that trajectories
provide useful information beyond the generated subtask.

\textbf{Autonomous driving.}
For NAVSIM~\citep{dauner2024navsim} and Waymo~\citep{xu2026wode2e} the policy
is a ReCogDrive-derived diffusion transformer~\citep{xiong2026recogdrive} that
reads the VLM's final-layer states over the prompt and the generated
plan through cross-attention, alongside ego history and vehicle state,
and outputs a continuous trajectory. The VLM is frozen and
generates a 2D+3D trajectory on each dataset's native grid; separate heads are
trained on Waymo, NAVSIM, and Alpamayo-derived data. RL then fine-tunes
the head alone with PDMS and RFS as rewards, so the plans are identical
in the SFT and RL rows.

As shown in Table~\ref{tab:navsim_waymo}, VFlash achieves competitive
driving performance with a frozen planning VLM and a small action head.
With supervised training, it reaches 87.1 PDMS on NAVSIM, above
ReCogDrive (86.8), and 7.79 RFS on Waymo, comparable to
Qwen-Drive-1.0-SFT (7.78). RL on the action head raises these scores
to 90.7 PDMS and 7.86 RFS, matching Qwen-Drive-1.0-RL on NAVSIM
and coming within 0.05 RFS of the best listed Waymo result.
These gains require no updates to the VLM or its generated plans.
Together, the results show that VFlash supports competitive
plan-conditioned driving while substantially reducing decoding
latency (Section~\ref{sec:throughput_analysis}).

\begin{table*}[t]
\centering
\captionsetup{font=small,skip=4pt}
\caption{
    \textbf{Autonomous driving results} on NAVSIM (left)
    and the Waymo test split (right).
    RL indicates reinforcement learning;
    $\uparrow$ / $\downarrow$ denote higher / lower is better.
}
\label{tab:navsim_waymo}

\begin{minipage}[t]{0.56\textwidth}
\vspace{0pt}
\centering
\small
\renewcommand{\arraystretch}{1.0}
\setlength{\tabcolsep}{5.2pt}

\resizebox{\linewidth}{!}{%
\begin{tabular}{lccccccc}
\toprule
\textbf{Method}
    & \textbf{RL}
    & \textbf{NC}$\uparrow$
    & \textbf{DAC}$\uparrow$
    & \textbf{EP}$\uparrow$
    & \textbf{TTC}$\uparrow$
    & \textbf{Comf.}$\uparrow$
    & \textbf{PDMS}$\uparrow$ \\
\midrule

Hydra-MDP~\citeyearpar{li2024hydra}
    & -- & 98.3 & 96.0 & 78.7 & 94.6 & 100.0 & 86.5 \\
DiffusionDrive~\citeyearpar{liao2025diffusiondrive}
    & -- & 98.2 & 96.2 & 82.2 & 94.7 & 100.0 & 88.1 \\
ReCogDrive~\citeyearpar{xiong2026recogdrive}
    & -- & 98.3 & 95.1 & 81.1 & 94.3 & 100.0 & 86.8 \\
\midrule

ReCogDrive~\citeyearpar{xiong2026recogdrive}
    & $\checkmark$ & 98.2 & 97.8 & 83.5 & 95.2 & 99.8 & 89.6 \\
SpanVLA~\citeyearpar{zhou2026spanvla}
    & $\checkmark$ & 99.1 & 97.1 & 86.3 & 95.2 & 100.0 & 90.3 \\
ExploreVLA~\citeyearpar{sheng2026explorevla}
    & $\checkmark$ & 98.8 & 98.4 & 83.5 & 96.5 & 99.9 & 90.4 \\
EponaV2~\citeyearpar{xu2026eponav2}
    & $\checkmark$ & 98.6 & 97.9 & 84.8 & 95.7 & 100.0 & 90.4 \\
\midrule

Qwen-Drive-1.0-SFT~\citeyearpar{zhou2026qwen}
    & -- & 98.4 & 96.6 & 82.4 & 94.7 & 100.0 & 88.2 \\
Qwen-Drive-1.0-RL~\citeyearpar{zhou2026qwen}
    & $\checkmark$ & 98.6 & 98.2 & 84.8 & 95.9 & 100.0 & 90.7 \\
\midrule

\rowcolor{modelrowcolor}
\textbf{VFlash-SFT}
    & -- & 98.1 & 95.2 & 81.7 & 94.3 & 100.0 & 87.1 \\
\rowcolor{modelrowcolor}
\textbf{VFlash-RL}
    & $\checkmark$ & 98.3 & 97.8 & 85.9 & 95.4 & 100.0 & 90.7 \\
\bottomrule
\end{tabular}%
}
\end{minipage}\hfill%
\begin{minipage}[t]{0.42\textwidth}
\vspace{0pt}
\centering
\small
\renewcommand{\arraystretch}{1.0}
\setlength{\tabcolsep}{4pt}
\resizebox{\linewidth}{!}{%
\begin{tabular}{lcccc}
\toprule
\textbf{Method}
    & \textbf{RL}
    & \textbf{ADE}$_{3\mathrm{s}}\downarrow$
    & \textbf{ADE}$_{5\mathrm{s}}\downarrow$
    & \textbf{RFS}$\uparrow$ \\
\midrule

DiffusionLTF~\citeyearpar{nguyen2025open}
    & -- & 1.36 & 2.89 & 7.72 \\
dVLM-AD~\citeyearpar{ma2026dvlm}
    & -- & 1.29 & 3.02 & 7.63 \\
HMVLM~\citeyearpar{wang2025hmvlm}
    & -- & 1.33 & 3.07 & 7.74 \\
MindVLA-U1~\citeyearpar{huang2026mindvla}
    & -- & 1.16 & 2.67 & 7.77 \\
Fast-dDrive~\citeyearpar{zhang2026fast}
    & -- & 1.25 & 2.91 & 7.82 \\
NoRD~\citeyearpar{rawal2026nord}
    & $\checkmark$ & 1.25 & -- & 7.71 \\
MindVLA-U1~\citeyearpar{huang2026mindvla}
    & $\checkmark$ & 1.09 & 2.66 & 7.87 \\
\midrule

Qwen-Drive-1.0-SFT~\citeyearpar{zhou2026qwen}
    & -- & 1.19 & 2.65 & 7.78 \\
Qwen-Drive-1.0-RL~\citeyearpar{zhou2026qwen}
    & $\checkmark$ & 1.19 & 2.67 & 7.91 \\
\midrule

\rowcolor{modelrowcolor}
\textbf{VFlash-SFT}
    & -- & 1.23 & 2.71 & 7.79 \\
\rowcolor{modelrowcolor}
\textbf{VFlash-RL}
    & $\checkmark$ & 1.22 & 2.70 & 7.86 \\
\bottomrule
\end{tabular}%
}
\end{minipage}

\end{table*}

\subsection{Throughput analysis}
\label{sec:throughput_analysis}

\begin{figure}[tbh]
    \centering
    \includegraphics[width=\linewidth]{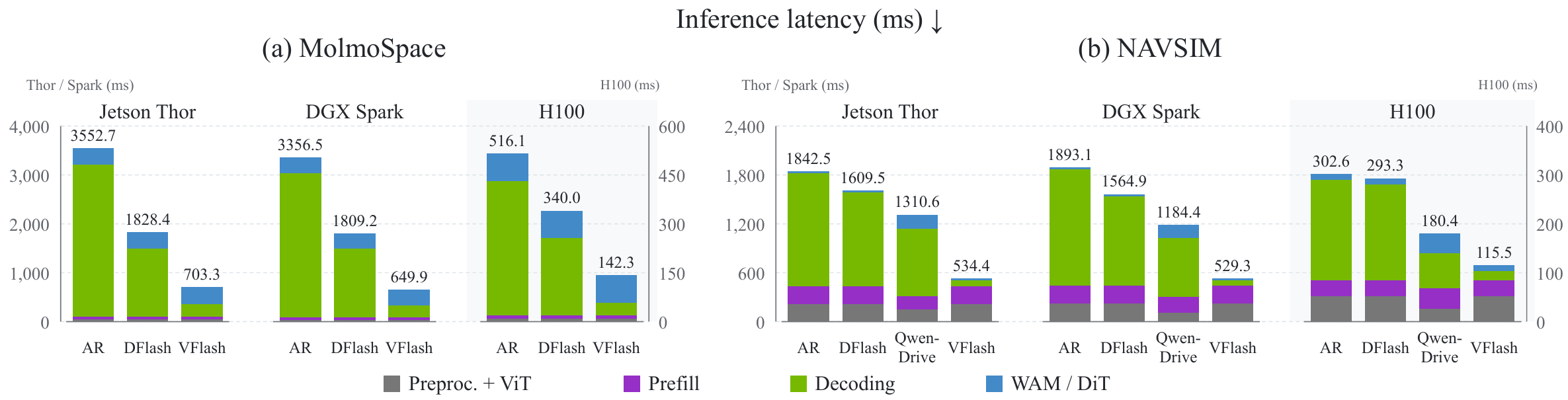}
    \captionsetup{font=small,skip=4pt}
    \caption{
    \textbf{VFlash} latency across different downstream tasks on Jetson Thor, DGX Spark and H100.
    }
    \label{fig:latency}
\end{figure}

Figure~\ref{fig:latency} splits per-step latency into visual encoding, prefill, decoding, and the action model at batch size one on Jetson Thor, DGX Spark, and H100. Thor is an embedded platform for robots and vehicles and Spark a desktop-class system, typical on-device hardware for the System-1 visual planning of Section~\ref{sec:introduction}; H100 is the server reference. Appendix~\ref{app:latency_eval} gives the timing boundary and the full table. Across devices, VFlash cuts per-step latency 5.1--5.2$\times$ on MolmoSpaces on Thor and Spark and 3.6$\times$ on H100, and 3.4--3.6$\times$ and 2.6$\times$ on NAVSIM; on Waymo (Figure~\ref{fig:waymo_latency}) the H100 reduction is 4.5$\times$. The gain comes almost entirely from decoding: under AR it takes 75--87\% of each step on Thor, and VFlash cuts it 12--42$\times$ while the other stages are unchanged.

\textbf{AR decoding} runs one 4B VLM pass per token (35--39~ms on Thor, 33--40~ms on Spark, 4.7--6.0~ms on H100), so decoding time grows roughly linearly with token count. \textbf{Drafting} predicts the remaining tokens of a 36--84-token primitive in one pass, taking 13--23~ms on Thor and 1--3~ms on H100. \textbf{Verification} still needs a target pass per block, so its benefit depends on the tokens committed per iteration. On MolmoSpaces, DFlash commits 5.5 tokens per iteration on subtask text but only 2.2 within trajectory blocks: a coordinate one bin off the target's choice fails Eq.~\ref{eq:verified_acceptance} even with small geometric error. DFlash thus only about halves decoding time. \textbf{VFlash} skips token-level verification for visual primitives, using one drafter pass and one target pass per primitive; the target pass updates the cache and extracts hidden states for the action head, amortized over the block. On Thor, decoding takes 72--78~ms as the driving primitive grows from 36 to 84 tokens, so denser plans add little latency over this range.

\section{Related Work}
\textbf{Embodied VLMs and visual plans.}
Embodied VLMs are increasingly used as high-level modules that turn
visual observations and language instructions into plans, affordances,
and trajectories~\citep{ji2025robobrain,baairobobrainteam2025robobrain20technicalreport,team2025geminirobotics,embodiedr1,rynnbrain}.
Spatial prediction is central to these models: points and boxes localize objects or
affordances~\citep{robopoint,roborefer,deitke2024molmo,rexomni}, while
traces describe intended or predicted motion~\citep{robotracer,wen2023atm,zhang2026molmomotion}.
These visual plans can condition manipulation
policies~\citep{rttrajectory,hamster,lohomanip,actionsketcher,huang2025vlmtdp}
or be fed back as prompts~\citep{zheng2024tracevla}. Other work reasons
spatially before acting~\citep{molmoact,fang2026molmoact2actionreasoningmodels,guo2026avp},
predicts future states~\citep{kim2026cosmospolicy,ye2026dreamzero}, or
plans driving trajectories from scene
understanding~\citep{drivevlm,wang2025omnidrive,nvidia2025alpamayo}.
While this literature asks what embodied models should predict, we ask
how fast such spatial predictions can be produced. Token-by-token
coordinate generation makes every extra waypoint, object, or dimension
an additional decoding step.

\textbf{Parallel and speculative decoding.}
Speculative decoding accelerates LLM inference by verifying multiple
drafted tokens in parallel~\citep{stern2018blockwise,leviathan2023fast,chen2023accelerating}.
Medusa, EAGLE, TIDAR, and DFlash explore different ways to produce these
drafts~\citep{cai2024medusa,li2024eagle,liu2026tidar,chen2026dflash}.
For visual outputs, LocateAnything decodes points and boxes as units,
but is limited to fixed-size 2D primitives~\citep{wang2026locateanything}.
Closest to this work, Spec-VLA accelerates OpenVLA by accepting draft action
tokens whose bins lie near the target's~\citep{wang2025specvla}. However, it still verifies each action token against the target and commits only $2.10$--$2.94$ tokens per pass. Based on our finding that, for visual trajectory prediction, one-pass drafts are more accurate than the autoregressive target, VFlash
instead drafts the entire visual primitive in one pass rather than by agreement with the target.

\textbf{Embodied and System-1 models.}
In a deployed perception-action loop, VLM latency directly limits how
quickly the agent can react. Prior efficiency works mainly accelerate
the action execution pipeline, for example via action compression, parallel
action chunks, or flow matching~\citep{pertsch2025fastefficientactiontokenization,openvlaoft,song2025pdvla,pi0}.
Prior VLM-based planning, however, still often rely on highly autoregressive token decoding.
System-1 models such as Jev make fast typed decisions that downstream
applications can use directly~\citep{almeida2026systemone}, and Visual Jev
answers many independent visual questions in parallel after one image
encoding~\citep{yu2026visualjev}. Our work can be seen as a relevant effort
that enables System-1 VLM spatial prediction. Within this context, our work
studies optimized token representation and decoding schemes systematically,
supporting both dedicated special tokens~\citep{chen2022pix2seq} and plain
numbers~\citep{chen2023shikra,bai2023qwenvlversatilevisionlanguagemodel,zhu2025internvl3}.

In summary, VFlash uses DFlash-style drafting without target-token agreement, extending fast and accurate visual plan generation from LocateAnything’s parallel box decoding to richer and more general visual primitives that are useful for downstream embodied applications. It also enables more efficient decoding parallelism via systematic studies, in contrast to relevant or concurrent works.

\section{Conclusion}
\label{sec:conclusion}

We presented VFlash, which uses the known structure of visual primitives
to replace sequential coordinate generation with parallel prediction.
Our results show that useful spatial predictions need not reproduce an
autoregressive model's output: direct drafts reduce robot trajectory
errors, while single-rollout timing estimates indicate approximately
$25\times$ faster generation for 3D object point forecasting.
Plan-conditioned manipulation further demonstrates how these predictions
support downstream action generation. Future work will examine whether
generating multiple complementary primitives from shared observations
improves control within a fixed latency budget.

\noindent\textbf{Acknowledgement.} The team would like to thank Pavlo Molchanov, Jindong Jiang, Yonggan Fu, Guillermo Siman, Andrew Tao, Karan Sapra, Pulkit Kumar, Patrick Liu and Yan Wang for the valuable discussions. The team would also like to give special thanks to the NVIDIA TSE Team for their exceptional work on the TensorRT deployment of this work.

\subsection*{AI use statement}

Generative AI tools were used to assist with manuscript drafting, language polishing, and structural editing, as well as with literature search and code development. 
AI-generated suggestions were used only as assistance and were reviewed, revised, and verified by the authors. All technical claims, experimental results, interpretations, and final written content were validated by the authors, who take full responsibility for the work.

\clearpage
\phantomsection
\addcontentsline{toc}{section}{References}
\begingroup
\small
\linespread{1.05}\selectfont
\setlength{\bibsep}{5pt}
\bibliographystyle{plainnat}
\bibliography{ref}
\endgroup

\clearpage
\appendix
\addtocontents{toc}{\protect\setcounter{tocdepth}{2}}
\begingroup
\renewcommand{\contentsname}{Appendix Contents}
\tableofcontents
\endgroup
\clearpage
\section{Visual Primitive Specifications}
\label{app:schemas}
\textbf{Special tokens and tool call.}
Our base VLM adds three groups of special tokens: the numbers
\texttt{<0>},\,\ldots,\,\texttt{<1000>}; the primitive delimiters listed
in Table~\ref{tab:primitive_spec}; and two markers, \texttt{<oof>} for
an out-of-frame image position and \texttt{<text\_mask>}, the mask token that occupies every position before the drafter predicts it. A primitive is requested by a tool call
that the target writes as ordinary text:
\begin{center}
\scriptsize \texttt{<tool\_call>\{"name":"atomic\_decoding","arguments":\{"block\_size":54\}\}</tool\_call>}
\end{center}
where \texttt{block\_size} counts every token of the block including its
start token. The start token that follows identifies the layout, and the
two together fix the role of every position before drafting. A block
whose length does not match any layout, or that exceeds the remaining
output budget, is rejected before drafting begins.

\begin{table}[!htbp]
\centering
\footnotesize
\setlength{\tabcolsep}{4pt}
\caption{\textbf{Primitive specifications for our base VLM.} $P$ is the
number of waypoints per arm or vehicle. Block length includes the start
token and all delimiters. Dual-arm blocks repeat the single-arm layout
for the left and then the right arm, with \texttt{left}/\texttt{right}
replacing \texttt{single} in the start tokens.}
\label{tab:primitive_spec}
\begin{tabular}{llc}
\toprule
Primitive & Layout & Length \\
\midrule
Bounding box & \texttt{<box>} $u_1 v_1 u_2 v_2$ \texttt{</box>} & $6$ \\
Robot 3D & \texttt{<robot\_single>} $(x\,y\,z)^{P}$ \texttt{</robot>} & $3P+2$ \\
Robot 2D+3D & \texttt{<robot\_single\_2d>} $(u\,v)^{P}$ \texttt{</robot>} & $5P+4$ \\
 & \texttt{<robot\_single>} $(x\,y\,z)^{P}$ \texttt{</robot>} & \\
Dual-arm 3D / 2D+3D & left-arm block, then right-arm block & $6P+4$ / $10P+8$ \\
Driving 3D & \texttt{<drive>} $(x\,y)^{P}$ \texttt{</drive>} & $2P+2$ \\
Driving 2D+3D & \texttt{<drive\_2d>} $(u\,v)^{P}$ \texttt{</drive>} & $4P+4$ \\
 & \texttt{<drive>} $(x\,y)^{P}$ \texttt{</drive>} & \\
\bottomrule
\end{tabular}
\end{table}

\textbf{Coordinate quantization.}
A coordinate $q\in[\ell,u]$ is mapped to a number token by
\begin{equation}
 k=\left\lfloor1000\frac{q-\ell}{u-\ell}+\frac{1}{2}\right\rfloor,
 \qquad \widehat q=\ell+\frac{k}{1000}(u-\ell),
 \label{eq:primitive_codec}
\end{equation}
rounding half up. Robot XYZ uses $[-5,5]$, $[-5,5]$, and $[0,10]$ meters
in the current primary camera's OpenCV frame (X right, Y down, Z
forward), giving a 1\,cm resolution. Driving uses ego-frame X forward in
$[-10,200]$ meters and Y left in $[-50,50]$ meters; height is omitted and
heading is left to the action head. Image coordinates are normalized to
$[0,1000]$: robot projections divide by the image dimension minus one,
boxes divide by width and height. These ranges are independent of the
action heads' own normalization (Appendix~\ref{app:driving_training}).

\textbf{Validity rules.}
A block is accepted only if every delimiter is in its declared position,
every coordinate position holds a number token, and the waypoint count
matches the tool call. A robot or driving waypoint whose projection falls
outside the image is written as a pair of \texttt{<oof>} markers in the
2D segment; its metric waypoint is kept. Boxes must satisfy $u_1<u_2$ and
$v_1<v_2$; the converter clips training boxes to the image and expands
zero-area boxes to one bin. Boxes within a category are generated one
block at a time, sorted lexicographically in training, and an empty
category is answered with the text \texttt{None}. Decoding enforces the
same rules by restricting each position to its legal tokens
(Appendix~\ref{app:decoding_algorithm}).

\textbf{Dense motion on a text-tokenized target.}
The MolmoMotion target keeps its own tokenizer and text format:
coordinates are signed millimeter integers relative to the first query
point at the current frame, and future frames are separated by a fixed
separator. One block covers one future frame of $P$ query points and has
$19P+3$ positions: the frame separator serves as the start token, two
positions hold the timestamp, and each point gets one position for its
index followed by three coordinates of a sign and five digits. Unused
high-order digits, and every position of a point the target would omit,
are filled with the target's mask token, so an omitted point is signaled
by a masked index rather than by a shorter frame. The completed block is
rendered back into the target's text by inserting its fixed punctuation;
the mask token never appears in the rendered answer. The target's
vocabulary is unchanged. Following MolmoMotion~\citep{zhang2026molmomotion}, we use $P=8$ and five-digit coordinates in all experiments (Appendix~\ref{app:upstream_training}).

\section{Training Details}
\label{app:training_decoding}
\label{app:upstream_training}

\textbf{Base VLM.}
The visual encoder is MoonViT-SO-400M~\citep{kimivl} with 27 layers of width 1,152,
14-pixel patches, and $2\times2$ spatial merging; a two-layer MLP
connects it to Qwen3-4B~\citep{qwen3}, whose decoder has 36 layers of width 2,560.
It is trained on image and video question answering, grounding,
pointing, and spatial reasoning data.
Every task-specific AR target in Section~\ref{sec:experiments} is fully
fine-tuned from this checkpoint. The dense-motion experiment instead
uses the released MolmoMotion-4B-H3-F30~\citep{zhang2026molmomotion} target without fine-tuning.

\textbf{Drafter architecture.}
Following DFlash~\citep{chen2026dflash}, the drafter has five
full-width Qwen3 transformer layers with 32 query heads, eight
key/value heads, 128-dimensional heads, and a 9,728-dimensional
feed-forward layer. Target layers 1, 9, 17, 25, and 33 (zero-based)
provide context features; their concatenated 12,800-dimensional states
are projected to 2,560 dimensions and RMS-normalized. The drafter
shares the frozen target's input embedding and output projection and
has 537M trainable parameters. Positions inside a block attend to each
other and to the target context before the block; target features are
computed online with gradients disabled.

\textbf{Loss weights.}
We minimize Eq.~\ref{eq:primitive_training} over the annotated
sequence, with the target encoding the clean sequence and the drafter
reconstructing the masked positions. Primitive positions have unit
weight. Ordinary-text positions decay as $w_j=\exp(-(j-1)/7)$ with
distance from the anchor, following DFlash. Out-of-frame labels are
downweighted by 0.1 in both the target's fine-tuning loss and the
drafter loss. Anchors and padding are excluded. No soft-logit
distillation or hidden-state regression is used.

\textbf{Block sampling and packing.}
Each annotated primitive contributes one complete block. The 2D,
2D+3D, and wrist-camera drafters additionally sample up to 512
ordinary-text anchors per example, shared across assistant turns, so
that one drafter serves both the text and the primitive path; text
blocks stop at primitive boundaries. The 3D-only and box drafters
train on primitive blocks only. Packed examples keep independent
position indices and attention masks: each block sees only its own
positions and the target positions of its own example before its
anchor, so schemas and lengths mix freely in a batch without leaking
labels across blocks or examples.

\begin{table}[!htbp]
\centering
\footnotesize
\setlength{\tabcolsep}{5pt}
\caption{\textbf{Task adaptation budgets.} AR columns describe full
target fine-tuning; draft columns describe drafter training on the
frozen target. Batch is the number of examples per draft update.}
\label{tab:upstream_recipes}
\begin{tabular}{lccccc}
\toprule
Task & AR updates & AR peak LR & Draft updates & Draft peak LR & Draft batch \\
\midrule
2D trajectories & 10,800 & $2{\times}10^{-6}$ & 7,142 & $2{\times}10^{-4}$ & 64 \\
Robot 3D-only & 10,317 & $2{\times}10^{-6}$ & 30,562 & $6{\times}10^{-4}$ & 512 \\
Robot 2D+3D & 11,207 & $2{\times}10^{-6}$ & 30,562 & $6{\times}10^{-4}$ & 512 \\
Robot 2D+3D, wrist camera & 26,245 & $2{\times}10^{-6}$ & 46,306 & $6{\times}10^{-4}$ & 512 \\
Driving 3D-only & 13,740 & $2{\times}10^{-6}$ & 4,028 & $6{\times}10^{-4}$ & 512 \\
Driving 2D+3D & 13,740 & $2{\times}10^{-6}$ & 4,028 & $6{\times}10^{-4}$ & 512 \\
Boxes & 3,000 & $2{\times}10^{-6}$ & 14,722 & $6{\times}10^{-4}$ & 256 \\
\bottomrule
\end{tabular}
\end{table}

\textbf{Data and optimization.}
Table~\ref{tab:upstream_recipes} lists the budgets. The 2D target uses
ShareRobot~\citep{ji2025robobrain} trajectories and affordances plus FSD~\citep{yuan2026seeing} direct and reasoning
annotations: 69,129 source records, 152,346 repeat-weighted examples per
pass, about two passes, two 8,192-token packs per update. Its drafter
trains for three passes with 16 examples per GPU on four GPUs and a
1,700-token limit per example.
The two primary-view robot targets share 5,215,780 repeat-weighted
examples from MolmoBot~\citep{deshpande2026molmobot}, AgiBot~\citep{agibot}, InternA1~\citep{tian2026interndata}, OXE-Trace~\citep{o2024open}, and DROID~\citep{droid},
serialized as 3D-only or joint 2D+3D trajectories; each AR run covers
about one pass with 16 packs of 32,768 tokens per update, and each
drafter trains for three passes with a 4,096-token limit per example.
The wrist-camera variant adds LIBERO~\citep{libero} two-camera observations and
additional views of the same robot data, for 7,902,844 weighted
examples; its target uses 32 packs of 16,384 tokens, its drafter an
8,192-token limit.
All runs use AdamW, BF16, zero weight decay, gradient clipping at 1.0,
and cosine decay to zero. Target warm-up is 200 updates for 2D
trajectories and 500 for robot 3D and boxes; drafter warm-up is 4\% of
the budget. Each drafter is paired with its own target and evaluated at
the final checkpoint.

\textbf{Boxes.}
The box target is trained on COCO train2017~\citep{lin2014microsoft}, CountAnything~\citep{lei2026count}, and
KubriCount~\citep{liu2026count} multi-turn conversations. Each COCO image is queried for its
present categories and up to ten sampled absent categories, with the
image supplied in the first turn only; three COCO variants differ in
their negative-category samples, and CountAnything and KubriCount use
positive categories. Repeating each source twice gives 1,256,254
weighted conversations per pass, with category outputs capped at 900
boxes. Post-training image identifiers do not overlap the evaluation
sets. The AR target trains for 3,000 updates with 16 packs of up to
32,768 tokens per update. The drafter trains for three passes with 16
conversations per GPU on 16 GPUs and a 262,144-token pack limit,
supervising every six-token box block with uniform weights; a mixed
text-and-box variant with up to 512 text anchors was also trained, but
the reported results use the box-only drafter.

\textbf{Dense motion.}
The MolmoMotion drafter is randomly initialized and trained with
uniform cross-entropy over the $19P+2$ predicted positions of each
frame block, mask tokens included. It uses the same five-layer
architecture and the frozen target's embedding and output weights. The
mixture contains 241,692 entries from EgoDex~\citep{egodex}, HD-EPIC~\citep{perrett2025hd}, Stereo4D~\citep{jin2025stereo4d},
Xperience~\citep{xperience_10m}, and YouTube-VIS~\citep{Yang2019vis}, sampled in proportion to the square root
of each source's size. Training windows use eight points, up to 30
future frames, and one or three history frames with equal probability;
frames whose coordinates exceed five digits are excluded. Training runs
34,563 updates over 12.7M sampled windows, each update using four
6,144-token packs per GPU on 32 H100 GPUs. AdamW uses learning rate
$2\times10^{-4}$, $(\beta_1,\beta_2)=(0.9,0.999)$, zero weight decay,
gradient clipping at 1.0, BF16 computation with FP32 master weights,
and 4\% linear warm-up followed by cosine decay to zero.
Table~\ref{tab:molmomotion} uses the final checkpoint.

\section{Decoding Details}
\label{app:decoding_algorithm}

Algorithm~\ref{alg:vflash} summarizes one generation. Text follows the
DFlash procedure and primitives follow VFlash; the two share the
drafter, the target, and the KV cache.

\begin{algorithm}[!htbp]
\caption{Flash primitive decoding for one response}
\label{alg:vflash}
\begin{algorithmic}[1]
\Require prompt $x$, target $\theta$, drafter $\phi$, block size $K=16$, budget $N$
\State prefill $\theta$ on $x$; $y\gets[\,]$
\While{$|y|<N$ and $y$ does not end with EOS}
  \If{$y$ ends with a primitive-start token $a_0$ of schema $s$}
    \State $B\gets$ block length declared by the preceding tool call
    \State $\hat a\gets\phi(H_\theta,\,a_0,\,\mathrm{MASK}^{B-1})$, one pass, each position restricted to its legal tokens
    \If{$A_s(\hat a)=1$} \Comment{schema valid; boxes also pass the confidence guard}
      \State append $\hat a$ to $y$; one target pass over $\hat a$ updates the cache and emits the next token
    \Else
      \State keep the first $r$ positions; regenerate the rest with constrained target AR
    \EndIf
  \Else
    \State $\hat a\gets\phi(H_\theta,\,y_{-1},\,\mathrm{MASK}^{K-1})$ \Comment{ordinary text}
    \State one target pass over $\hat a$; accept the longest agreeing prefix plus the target's correction
    \State truncate after a primitive-start token if one was accepted
  \EndIf
\EndWhile
\end{algorithmic}
\end{algorithm}

\textbf{Text and routing.}
After multimodal prefill, ordinary text is generated in DFlash blocks
of 16 positions: one accepted anchor and up to 15 proposals. The target
verifies the block in one pass and keeps the longest matching prefix
plus its own next token. The tool call is generated on this path. When
an accepted prefix reaches a primitive-start token, any speculative
tokens after it are discarded and the declared schema and block length
take over. A block whose length exceeds the remaining budget is
rejected before drafting.

\textbf{Primitive prediction and acceptance.}
The drafter processes the start token and all masked positions in one
pass. Delimiter positions are fixed, coordinate positions are restricted
to number tokens, and coupled rules such as the pairing of out-of-frame
markers and the ordering of box corners are resolved by a small
selection loop over the logits already computed, with no further
network evaluation. The target then processes the completed block once
to update its cache, supply context features for the next block, and
predict the continuation. A schema-valid block is accepted in full
without token-level agreement. If validation fails, the target cache is
rolled back to the anchor and the block is regenerated by constrained
target AR from the first invalid position ($r=0$ for trajectories); the
drafter cache is reset afterwards. The DFlash baseline in
Section~\ref{sec:throughput_analysis} instead applies 16-position blocks
and longest-prefix acceptance throughout the response, including inside
primitives.

\textbf{Dense motion continuation.}
For the MolmoMotion target the drafter fills the $19P+2$ positions of
one frame.
The rendered frame receives one target pass, which supplies the next
frame's features and chooses between the frame separator and the
closing delimiter; the target writes the answer's opening and closing
text autoregressively. Missing or non-increasing timestamps are set to
the previous timestamp plus one, and a frame whose points are all
masked keeps its highest-margin point. Malformed frames fall back to
target AR. Generation continues for at most 30 frames, so parallelism
is across the points and coordinates of one frame while frames remain
sequential.

\section{Downstream Action Models}
\label{app:action_models}
\subsection{MolmoSpaces}
\label{app:molmospaces_protocol}

\textbf{Plan conditioning.}
The manipulation policy is the world action model (WAM) of
MolmoSpaces~\citep{kim2026molmospaces}, obtained by post-training
Cosmos-Predict2.5~\citep{nvidia2026worldsimulation} for robot control. 
Unless otherwise specified, MolmoSpaces uses the primary-view 2D+3D robot target–drafter pair described in Appendix~\ref{app:upstream_training}.
We replace its text-encoder context with states from the frozen 4B
planning VLM. For every retained position, each of the VLM's 36 layer
outputs is standardized along its 2,560 channels and the results are
concatenated; a newly initialized linear projection followed by GELU
maps this vector to the WAM's context width. The resulting 512 context
positions enter the WAM through its existing text cross-attention path.
The trajectory reaches the WAM only through these states; there is no
separate numeric trajectory encoder. The VLM stays frozen throughout.

\textbf{Post-training.}
All five configurations start from the same pretrained WAM checkpoint
and are evaluated after 100,000 post-training updates. Post-training
uses AdamW with learning rate $10^{-4}$, weight decay 0.01,
gradient-norm clipping at 1.0, BF16, and global batch size 512 (32
GPUs, four examples per device, four accumulation steps); the schedule
warms up for 15,000 steps and stays at its peak through the evaluated
checkpoint, for 51.2 million samples per model. Training uses seed 24,
and every configuration is evaluated at the same update count rather
than at its own best checkpoint. The post-training data are 172.5
million frames of simulated Franka demonstrations in six task splits
covering picking, placing, and opening

\textbf{Execution and scoring.}
The WAM executes 20-action chunks. A new plan is generated synchronously every three chunks, or 60 control steps; the simulator pauses during inference.
Evaluation uses seed 42 and all 3,915 benchmark episodes: 1,000 each
for Pick, Pick \& Place, and Open, and all 915 Close episodes. Horizons
are 455 control steps for Pick, Open, and Close and 682 for Pick \&
Place. An episode succeeds if the goal is reached at any step; episodes
run to the horizon. Suite success is aggregated before rounding, and
the mean is the unweighted average of the four suites. Each row is one
training seed and one evaluation seed.

\subsection{LIBERO}
\label{app:libero_protocol}

\textbf{Action head.}
The LIBERO policies use the PIv3 flow-matching action head of
StarVLA~\citep{community2026starvla}. The LIBERO experiments use the wrist-camera-trained 2D+3D robot target–drafter pair described in Appendix~\ref{app:upstream_training}.
A LayerNorm and linear projection
map each of the frozen VLM's 36 hidden-state layers from 2,560 to 1,024
dimensions, and action-transformer layer $i$ cross-attends to projected
VLM layer $i$. The 36-layer head and projections contain about 634
million trainable parameters. Both primary and wrist images are rotated
by $180^\circ$, resized to $224\times224$, and included in the VLM
sequence; no proprioceptive input is used. The head predicts eight
seven-dimensional actions (end-effector increments and a gripper
command) with four flow-matching steps. VFlash generates a fresh subtask
and plan before every action chunk. The two context variants retain the
prompt through the generated subtask, or the prompt through the full
generated answer. The baseline is the same head on a frozen Qwen3-VL-4B
that encodes only the images and the instruction.

\textbf{Training and evaluation.}
All policies train on 273,463 demonstration steps from the same 1,693
episodes. The randomly initialized heads train for 80,000 AdamW updates
with batch size 128, learning rate $10^{-4}$, 5,000 warm-up steps,
cosine decay to $10^{-6}$, $(\beta_1,\beta_2)=(0.9,0.95)$, weight decay
$10^{-8}$, BF16, and seed 42. Evaluation uses seed 7.
LIBERO~\citep{libero} covers four suites with ten tasks and 50 initial
states each (2,000 episodes). LIBERO-Plus~\citep{liberoplus} evaluates
10,030 perturbed task instances once each. LIBERO-PRO~\citep{zhou2025liberopro} crosses the four
base suites with object, swap, language, and task perturbations (8,000
episodes); environment perturbations are excluded. The policies train
on the standard 40 LIBERO tasks and are evaluated unchanged on the two
perturbation benchmarks. Success follows each environment's goal
predicate. We report the unweighted mean of the success rates across its four task suites.

\subsection{Autonomous Driving}
\label{app:driving_training}

\subsubsection{Action Model Training}
\label{app:av_dit_training}

\textbf{Conditioning on the target VLM.}

The action head conditions on selected 2,560-dimensional final-layer states of the frozen target VLM. States at generated trajectory positions are obtained by processing the accepted primitive with the target VLM. We define conditioning sequence as prompt and plan. Variable-length BF16 states are projected to 384 dimensions, with padding masked in pooling and cross-attention. 

\textbf{Ego motion inputs.}
The head uses 16 history entries, each containing position and heading $(x,y,\psi)$, three availability indicators, and a relative timestamp. The resulting 112-dimensional vector is standardized using training-set statistics, with unavailable values zeroed and shorter histories left-padded. The 16-dimensional current-state vector contains four command indicators, four motion measurements, and eight availability indicators. Separate MLPs with 1,024 hidden units and LayerNorm map the history and current state to 384-dimensional embeddings.

\textbf{Trajectory head.}
The ReCogDrive-derived head alternates eight self-attention and eight cross-attention blocks, with eight heads of width 48, adaptive normalization, and SwiGLU feed-forward layers. Each noisy waypoint is embedded with a learned index and a sinusoidal encoding of its physical future time; the index table supports up to 64 waypoints. Concatenating the waypoint embedding with the broadcast history embedding and masked mean VLM context yields 1,152 channels, which are projected to 384. The current-state and diffusion-timestep embeddings condition the transformer blocks. A 512-dimensional output projection followed by an MLP with 1,024 hidden units predicts three noise components per waypoint, for $(x,y,\psi)$ on NAVSIM and $(x,y)$ on Waymo, where the heading channel is masked. NAVSIM and Waymo use the same architecture with separately trained weights.

\textbf{Diffusion objective and sampling.}
For normalized recorded future waypoints $\widetilde Y$, the head $\epsilon_\theta$ minimizes masked noise-prediction error:
\begin{equation}
 Y_s=\sqrt{\bar\alpha_s}\,\widetilde Y
       +\sqrt{1-\bar\alpha_s}\,\epsilon,
 \qquad
 \mathcal L_{\mathrm{drive}}=
 \mathbb E\!\left[
 \frac{\sum_{h,c}m_{hc}
   \bigl(\epsilon_{\theta,hc}(Y_s,s;Z,u,\tau)-\epsilon_{hc}\bigr)^2}
      {\sum_{h,c}m_{hc}}
 \right].
\end{equation}
Here $\epsilon\sim\mathcal N(0,I)$, $s$ is the diffusion timestep, $\bar\alpha_s$ is the cumulative noise schedule, $Z$ denotes VLM states, $u$ contains ego history and state, and $\tau$ contains future timestamps. The mask $m_{hc}$ is one for valid coordinates and zero otherwise; the heading channel is masked out on Waymo. The schedule has 100 training timesteps. Five deterministic DDIM steps~\citep{song2020denoising}, at indices 80, 60, 40, 20, and 0, produce the continuous trajectory. During SFT, only the action head and its input projections receive gradients.

\textbf{Normalization and training configuration.}
Each channel is mapped affinely to $[-1,1]$, $z_{\mathrm{norm}} = 2(z - z_{\min})/(z_{\max} - z_{\min}) - 1$, with bounds set per dataset. The NAVSIM head inherits the fixed ReCogDrive bounds: X in $[-1.57, 65.17]$\,m, Y in $[-19.68, 22.32]$\,m, and $\psi$ in $[-1.67, 1.86]$\,rad; targets are not clipped, while denoised predictions and the final normalized actions are clipped to $[-1,1]$. The Waymo head uses X in $[-46.98, 138.77]$\,m and Y in $[-18.14, 17.81]$\,m, fitted on training trajectories as the smaller of zero and the minimum to the 99th percentile for X and the 0.5th to 99.5th percentiles for Y, with no clipping. Both heads are initialized from scratch and trained for 200 epochs with AdamW at learning rate $10^{-4}$, weight decay $10^{-4}$, $(\beta_1,\beta_2)=(0.9,0.95)$, and gradient-norm clipping at 1.0; three warmup epochs are followed by cosine decay to $10^{-6}$. Training uses batch size 64 per GPU on eight GPUs with FP16 mixed precision.

\textbf{Reinforcement learning.}
After SFT, each head is fine-tuned with DiffGRPO~\citep{xiong2026recogdrive}, which treats the five-step DDIM chain as a policy rollout. For every training scene, the head samples eight trajectories, scores them with a dataset-specific reward, and updates with group-standardized advantages. A behavior-cloning loss with weight 0.1 regularizes the update. It is the negative log-likelihood under the current head of denoising chains sampled from the frozen SFT head. No KL penalty is used. Both heads use an initial learning rate of $10^{-5}$ with cosine decay and no warmup. On NAVSIM the reward is PDMS from the non-reactive simulator~\citep{dauner2024navsim} on the training split, with ego progress, time-to-collision within bound (TTC), and comfort weighted 10/5/2 during training against the official 5/5/2 used for evaluation. Training runs 10 epochs (6,650 updates) at batch size 16 per GPU on eight GPUs. The checkpoint is selected by PDMS on an internal validation split held out from navtrain by log (18,179 clips, Appendix~\ref{app:driving_protocols}). On Waymo the reward is RFS from the official implementation on the 479 validation scenes. Training runs 20 epochs (300 updates) at batch size 4 per GPU on eight GPUs, and the checkpoint is selected on the same 479 scenes, so the selection score is in-sample. These settings are shared by all layout and context configurations. The VLM, its plans, and the conditioning states stay fixed, so the SFT and RL rows of Table~\ref{tab:navsim_waymo} consume identical plans and differ only in head weights. Reported scores use the held-out test splits, which take no part in SFT, RL, or checkpoint selection.

\clearpage
\section{Latency Evaluation}
\label{app:latency_eval}
\subsection{Deployment and Configuration}
\label{app:latency_configuration}

\Needspace{22\baselineskip}
\begin{wrapfigure}[19]{r}{0.46\textwidth}
    \vspace{-16pt} 
    \centering
    \includegraphics[width=0.40\textwidth]{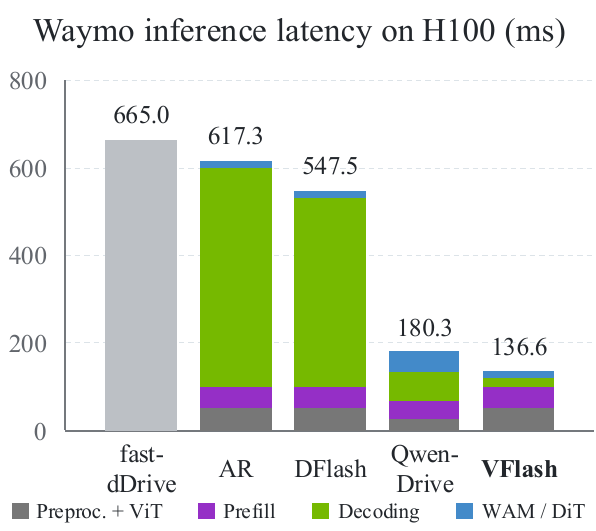}
    \captionsetup{font=small,skip=4pt}
    \caption{\textbf{Waymo planning latency on H100} (ms, batch size one). AR, DFlash, and VFlash share the VLM, primitive, and action head and differ only in the decoder. Qwen-Drive-1.0 is measured on the same hardware; the Fast-dDrive value is from its paper and its AR baseline is not the AR bar shown here.}
    \label{fig:waymo_latency}
\end{wrapfigure}

\textbf{Hardware and runtime.}
All measurements use batch size one and one request at a time on NVIDIA Jetson AGX Thor (SM110), DGX Spark (GB10, SM121), and a single H100 80\,GB (SM90). Our models run in a locally modified TensorRT-Edge-LLM\citep{nvidia2026tensorrt_edge_llm} C++ runtime: vision encoder, target VLM, and drafter are exported through ONNX and compiled into TensorRT engines, TensorRT selects kernels for the exported operators, and Edge-LLM plugins supply attention (CuTe~DSL FMHA for vision and prefill, XQA for decoding and speculative verification), cache updates, and positional encoding. Engines and kernel artifacts are built separately for each architecture. The stacks are CUDA~13.0 and TensorRT~10.15.2.7 on Thor, CUDA~13.0.88 and TensorRT~10.14.1.48 on Spark, and CUDA~12.9.86 and TensorRT~10.11.0.33 on H100. All Thor runs use a modified Edge-LLM~0.9.0 tree; all Spark runs and the H100 MolmoSpaces runs use Edge-LLM~0.10.1; the H100 driving runs use the 0.9.0 tree with an SM90 attention bridge. Vision, target, and draft engines run in FP16 with FP16 KV caches; no weight or KV quantization is used.

\textbf{Decoders.}
AR performs target-only single-token decoding. DFlash drafts 16-token blocks and accepts the longest prefix that matches the target's greedy choice exactly, plus the correction token; numeric coordinates receive no tolerance. VFlash decodes text with the same verified blocks, then drafts the whole primitive in one unverified block (54 tokens for MolmoSpaces, 36 for NAVSIM, 84 for Waymo) and runs one cached target prefill over the accepted block, the EOS token, and end-of-text padding to produce the hidden states the action model consumes. This prefill is counted as decoding. For MolmoSpaces the prefill retains 512 non-image positions from all 36 target layers; for driving it exports the final layer only, padded to 2,200 positions on NAVSIM. In both pipelines the fixed tool-call JSON is supplied rather than generated; on MolmoSpaces it is committed inside the text verification pass, and on driving the think-and-tool prefix precedes generation, so decoding covers exactly the primitive. All decoders use the same supplied tool-call JSON and the same primitive length. AR and DFlash reuse the target states computed during generation and need no separate feature pass.

\textbf{Action models.}
The MolmoSpaces WAM runs one WAN encoder call and four DiT calls (joint, action, joint, action) inside a CUDA graph, with BF16 DiT weights and FP16 attention through a custom head-dimension-128 CuTe~DSL kernel on Thor and Spark; the H100 baseline uses native TensorRT attention. The ReCogDrive head is a BF16 TensorRT engine containing feature encoding and the DDIM sampler. Latency profiles unroll ten DDIM steps to match the ten-step Qwen-Drive planner; the quality results of Table~\ref{tab:navsim_waymo} use five steps (Appendix~\ref{app:driving_protocols}).

\textbf{Qwen-Drive-1.0.}
We run the released checkpoint in its reasoning planning mode with greedy decoding. On Thor and Spark it uses a customized Edge-LLM~0.9.0 runtime with CuTe~DSL Gated DeltaNet kernels for its 24 linear-attention layers and FMHA/XQA for its eight full-attention layers; on H100 it uses the TensorRT-LLM backend with FlashInfer kernels. Its 1.04B flow-matching planner is a separate TensorRT engine with BF16 weights, FP16 attention, and ten Euler steps in a CUDA graph. Inputs are one 1600$\times$896 front image on NAVSIM (1,400 visual tokens; 1600$\times$800 in the Thor runs) and three 800$\times$896 images on Waymo (2,100 tokens). The Fast-dDrive value in Figure~\ref{fig:waymo_latency} is taken from its paper and was not measured here.

\textbf{Timing protocol.}
Each request starts from fresh vision computation and model state. CUDA events bracket each stage (preprocessing and vision encoding, prefill, decoding, action model) and are summed over all calls within a request; these intervals include host launch gaps and are not kernel-only durations. Model loading, engine construction, graph capture, warmup, image loading and CPU preprocessing outside the instrumented modules, and output detokenization are excluded. Repeat counts differ by run: the Thor MolmoSpaces pipeline uses 24 records and ten passes after five warmups; the Spark and H100 robotics runs one pass over the same 24 records after five warmups; driving runs five records and ten passes after five warmups; the Qwen-Drive H100 run five scenes twice after five warmups. Reported per-step latency is the sum of stage means. Within each device and task, the vision, prefill, and action-model means are shared across decoders; decoding is profiled separately. The MolmoSpaces AR decoding time comes from an earlier standalone profile with an output-only target engine, whereas the VFlash decoding time includes the all-layer feature completion described above; the driving decoders share one target-engine configuration.

\subsection{Profiling Results}
\label{app:latency}

Table~\ref{tab:latency_full} lists the stage means behind
Figures~\ref{fig:latency} and~\ref{fig:waymo_latency}, and
Table~\ref{tab:decoding_stats} the decoding statistics behind the
per-pass numbers of Section~\ref{sec:throughput_analysis}.

\begin{table}[p]
\centering
\linespread{1.0}\selectfont
\footnotesize
\setlength{\tabcolsep}{5pt}
\captionsetup{font=footnotesize,skip=4pt}
\caption{\textbf{Per-step latency by stage (ms).} Batch size one, FP16
VLM engines. Action is WAN encoder plus DiT for MolmoSpaces and the
ReCogDrive DiT (ten DDIM steps) or the Qwen-Drive planner for driving.
Within each task and device, vision, prefill, and action means are
shared across AR, DFlash, and VFlash and shown once; decoding is
profiled separately and for VFlash includes the target prefill over the
accepted primitive. Total is the sum of stage means. Qwen-Drive-1.0 is
measured end to end on its own stack (Appendix~\ref{app:latency_configuration});
Fast-dDrive is reported by its paper without a stage split.}
\label{tab:latency_full}
\begin{tabular}{llrrrrrr}
\toprule
Device & Method & Prepro. & ViT & Prefill & Decoding & Action & Total \\
\midrule
\multicolumn{8}{l}{\textit{MolmoSpaces}} \\
\addlinespace[2pt]
\multirow{3}{*}{Thor}  & AR     & \multirow{3}{*}{19.1} & \multirow{3}{*}{24.4} & \multirow{3}{*}{61.8} & 3,107.6 & \multirow{3}{*}{339.8} & 3,552.7 \\
                       & DFlash &  &  &  & 1,383.3 &  & 1,828.4 \\
                       & VFlash &  &  &  &   258.2 &  &   703.3 \\
\addlinespace[3pt]
\multirow{3}{*}{Spark} & AR     & \multirow{3}{*}{18.2} & \multirow{3}{*}{19.0} & \multirow{3}{*}{53.4} & 2,950.9 & \multirow{3}{*}{315.0} & 3,356.5 \\
                       & DFlash &  &  &  & 1,403.6 &  & 1,809.2 \\
                       & VFlash &  &  &  &   244.3 &  &   649.9 \\
\addlinespace[3pt]
\multirow{3}{*}{H100}  & AR     & \multirow{3}{*}{4.0} & \multirow{3}{*}{5.8} & \multirow{3}{*}{10.1} &   411.8 & \multirow{3}{*}{84.4} &   516.1 \\
                       & DFlash &  &  &  &   235.7 &  &   340.0 \\
                       & VFlash &  &  &  &    38.0 &  &   142.3 \\
\midrule
\multicolumn{8}{l}{\textit{NAVSIM}} \\
\addlinespace[2pt]
\multirow{3}{*}{Thor}  & AR     & \multirow{3}{*}{18.6} & \multirow{3}{*}{195.0} & \multirow{3}{*}{223.9} & 1,380.0 & \multirow{3}{*}{25.0} & 1,842.5 \\
                       & DFlash &  &  &  & 1,147.0 &  & 1,609.5 \\
                       & VFlash &  &  &  &    71.9 &  &   534.4 \\
                       & Qwen-Drive-1.0 & 12.2 & 139.0 & 160.0 & 828.4 & 171.0 & 1,310.6 \\
\addlinespace[3pt]
\multirow{3}{*}{Spark} & AR     & \multirow{3}{*}{16.0} & \multirow{3}{*}{205.3} & \multirow{3}{*}{219.0} & 1,428.2 & \multirow{3}{*}{24.6} & 1,893.1 \\
                       & DFlash &  &  &  & 1,100.0 &  & 1,564.9 \\
                       & VFlash &  &  &  &    64.4 &  &   529.3 \\
                       & Qwen-Drive-1.0 & 12.0 & 100.8 & 192.0 & 720.9 & 158.6 & 1,184.4 \\
\addlinespace[3pt]
\multirow{3}{*}{H100}  & AR     & \multirow{3}{*}{5.0} & \multirow{3}{*}{47.2} & \multirow{3}{*}{31.8} &   206.3 & \multirow{3}{*}{12.3} &   302.6 \\
                       & DFlash &  &  &  &   197.0 &  &   293.3 \\
                       & VFlash &  &  &  &    19.2 &  &   115.5 \\
                       & Qwen-Drive-1.0 &  5.0 &  21.5 &  42.0 &  71.9 &  40.0 &   180.4 \\
\midrule
\multicolumn{8}{l}{\textit{Waymo}} \\
\addlinespace[2pt]
\multirow{3}{*}{Thor}  & AR     & \multirow{3}{*}{26.0} & \multirow{3}{*}{277.8} & \multirow{3}{*}{359.9} & 3,254.0 & \multirow{3}{*}{32.0} & 3,949.7 \\
                       & DFlash &  &  &  & 2,431.6 &  & 3,127.3 \\
                       & VFlash &  &  &  &    78.3 &  &   774.0 \\
                       & Qwen-Drive-1.0 & 13.5 & 170.1 & 252.5 & 749.7 & 195.5 & 1,381.3 \\
\addlinespace[3pt]
\multirow{3}{*}{Spark} & AR     & \multirow{3}{*}{26.0} & \multirow{3}{*}{240.5} & \multirow{3}{*}{333.1} & 3,390.0 & \multirow{3}{*}{31.2} & 4,020.8 \\
                       & DFlash &  &  &  & 2,804.2 &  & 3,435.0 \\
                       & VFlash &  &  &  &    69.5 &  &   700.3 \\
                       & Qwen-Drive-1.0 & 13.5 & 123.5 & 280.8 & 651.2 & 180.2 & 1,249.2 \\
\addlinespace[3pt]
\multirow{3}{*}{H100}  & AR     & \multirow{3}{*}{6.0} & \multirow{3}{*}{45.5} & \multirow{3}{*}{45.4} &   501.8 & \multirow{3}{*}{18.6} &   617.3 \\
                       & DFlash &  &  &  &   432.0 &  &   547.5 \\
                       & VFlash &  &  &  &    21.1 &  &   136.6 \\
                       & Qwen-Drive-1.0 &  5.0 &  21.0 &  41.0 &  65.5 &  47.8 &   180.3 \\
                       & Fast-dDrive & -- & -- & -- & -- & -- & 665.0 \\
\bottomrule
\end{tabular}

\bigskip
\caption{\textbf{Decoding statistics on MolmoSpaces (Thor).} Tokens
committed per target pass, passes per request, and per-pass cost of the
drafter and target. The output is 88.75 tokens on average: about 36 of
text and tool call, then the 54-token primitive. VFlash's text stage
uses the same verified blocks as DFlash; the fused tool-call JSON is
excluded from its committed count. DFlash's decoding time in
Table~\ref{tab:latency_full} additionally includes a final padding
forward that completes the feature sequence (58.5\,ms on Thor, 54.5 on
Spark, 11.2 on H100); for VFlash this work is folded into the target
prefill over the primitive.}
\label{tab:decoding_stats}
\begin{tabular}{llrrrr}
\toprule
Decoder & Segment & Tokens/pass & Passes & Drafter ms & Target ms \\
\midrule
AR     & all         & 1.0  & 88.75 & --   & 35.0 \\
DFlash & all         & 3.0  & 29.5  & 9.1  & 35.7 \\
       & text        & 5.5  &       &      &      \\
       & trajectory  & 2.2  &       &      &      \\
VFlash & text        & 5.5  & 3.25  & 11.8 & 42.5 \\
       & primitive   & 53   & 1     & 12.9 & 68.4 \\
\bottomrule
\end{tabular}
\end{table}

\section{Evaluation Protocols}
\label{app:protocols}

\subsection{2D Trajectories}
\label{app:open_loop_2d_protocol}

We evaluate 200 ShareRobot-Bench queries and 300 VABench-V queries with
the same 4B target and its drafter. Coordinates are on the 0--1,000
image scale. Predictions are linearly interpolated by waypoint index to
the reference point count before computing discrete Fr\'echet distance,
symmetric Hausdorff distance, and RMSE, the square root of the mean
squared Euclidean waypoint error; Table~\ref{tab:open_loop_2d} divides
each distance by 1,000. All AR and VFlash outputs parse and enter the
means.

\subsection{3D Robot Trajectories}
\label{app:robot_3d_protocol}

\textbf{Inputs and references.}
Every decoder is evaluated on the same 1,000 InternA1, 999 AgiBot, and
4,221 MolmoBot records. InternA1 uses the current image; AgiBot
contains 499 image and 500 video records, MolmoBot 2,757 and 1,464.
Video records sample six frames from the preceding three seconds and
exclude the current frame. Each record supplies only the primary
camera, including for the model trained with wrist-camera views.
References contain ten waypoints per gripper, uniformly resampled by
3D arc length over a nominal five-second window that may end early at
an episode or action boundary; they describe path geometry rather than
fixed-time control samples. Coordinates are in meters in the current
primary camera frame and lie on a 1\,cm grid inherited from the
annotation export.

\textbf{Metrics.}
The scorer requires the expected arm identities and exactly ten points
per arm, without interpolation or repair; all reported configurations
return valid trajectories for every record. ADE averages Euclidean
waypoint error and FDE the final-waypoint error over the evaluated arms.
SR@$\tau$ is the fraction of waypoints with error below $\tau$, averaged
within a record and then across records, for $\tau=5$ and 10\,cm. An
arm is active when the sum of its XYZ standard deviations is at least
5\,mm; if any reference arm is active, only active arms are scored.
InternA1 evaluation episodes are absent from training; MolmoBot uses
five held-out validation configurations covering pick, place, and
opening tasks.

\subsection{Dense Motion}
\label{app:molmomotion_protocol}

\textbf{Inputs and selection.}
Each record gives three observed frames, the caption, and the observed
XYZ coordinates of eight query points, and asks for up to 30 future
frames. We use PointMotionBench with one fixed
point selection shared by all local evaluations: points must be visible
with finite coordinates in all three observed frames; at most 24 are
sampled per clip with seed zero and split into groups of eight,
discarding incomplete groups; HOT3D uses one fifth of its clips. This
gives 491/144/88 clips and 1,473/412/257 records for
HOT3D/WorldTrack/DAVIS. Frames are consecutive at the source rate, so
the horizon is 1.0\,s for HOT3D and WorldTrack and 1.25\,s for DAVIS;
shorter clips keep their available frames. WorldTrack uses the provided
world coordinates.

\textbf{Speed annotation.}
The $25\times$ in Table~\ref{tab:molmomotion} compares one
VFlash rollout with the target's AR rollout on the same selections,
both on one H100 at batch size one: 3.21/3.86/3.78\,s against
79.8/90.6/99.6\,s on HOT3D/\allowbreak{}WorldTrack/\allowbreak{}DAVIS, or
24.8/23.5/26.4$\times$.

\subsection{Box Generation}
\label{app:box_protocols}

\textbf{Queries and generation.}
Evaluation uses 4,952 COCO val2017 images and 200 Dense200 images. The
prompt asks for every instance of a given category in the primitive
format. COCO queries each ground-truth-present category as one turn of
a multi-turn conversation over the image; Dense200 has one category per
image. AR and VFlash share the image, query, and reference records,
decode greedily with neutral repetition penalty, and apply no resizing
before the model's image processor. The output budget is 512 new tokens
per COCO turn and 4,096 per Dense200 query; only complete boxes are
returned when the budget runs out. Boxes are mapped back to pixels and
merged per image.

\textbf{Direct and AR fallback.}
Direct accepts every schema-valid box. AR fallback adds the
spatial-confidence criterion of
LocateAnything~\citep{wang2026locateanything}: for each coordinate, the
numeric tokens among the five most probable full-vocabulary predictions
are examined, and the box is regenerated by constrained target AR if
this set is empty, or if the top numeric token has probability below
0.9 and at least two numeric candidates span more than 60 bins. Boxes
before the first uncertain coordinate are kept
(Eq.~\ref{eq:direct_acceptance}). On COCO a third consecutive identical
box ends the list, keeping the first two copies; Dense200 uses the
guard without this rule.

\textbf{Metrics.}
COCO F1 follows the inherited evaluator: every prediction receives
confidence 0.2, and per-category precision and recall are aggregated
into F1 at each IoU threshold; this is distinct from COCO AP. Dense200
greedily matches ground-truth boxes to unused predictions at each
$\tau\in\{0.50,0.55,\ldots,0.95\}$, averages per-image precision and
recall, and computes
$F_{1,\tau}=2\overline P_\tau\overline R_\tau/(\overline P_\tau+\overline R_\tau)$;
mean F1 averages the ten thresholds. Boxes per second (BPS) divides the
number of returned boxes by generation time on one H100 at batch size
one, where generation time covers all queries of an image including
prefill, target passes, and fallback, and excludes model loading.

\subsection{Autonomous Driving}
\label{app:driving_protocols}

\textbf{Splits and data separation.}
NAVSIM training uses navtrain, divided by log into 85,109 training and 18,179 internal validation clips; SFT and RL fit on the former, and checkpoints are selected on the latter by PDMS. Test scores use all 12,146 navtest scenes. Waymo SFT uses the official training split. The 479 validation scenes provide the RL reward and are used to select both SFT and RL checkpoints by RFS. Test scores come from the official test split. Neither test split enters SFT, RL, or checkpoint selection. The upstream VLM was fine-tuned on all of navtrain and on the Waymo training split. The NAVSIM internal validation clips are therefore held out from action-head fitting but not from VLM fine-tuning. The Waymo validation scenes and both test splits are excluded from VLM fine-tuning.

\textbf{Scoring.}
Each scene is evaluated once: the initial Gaussian noise is seeded from the clip identifier, and five deterministic DDIM steps produce the trajectory. On NAVSIM the head's eight $(x,y,\psi)$ waypoints at 2\,Hz are scored by the official non-reactive PDM simulator~\citep{dauner2024navsim} with the standard 5/5/2 weights for ego progress, time-to-collision within bound, and comfort. On Waymo the head's twenty $(x,y)$ waypoints at 4\,Hz are scored by the official RFS implementation, and ADE at 3\,s and 5\,s averages Euclidean position error over the horizon; both require a complete prediction through the horizon. 

\textbf{Compared methods.}
DiffusionLTF is the DiffusionDrive-based policy released with Open X-AV~\citep{nguyen2025open}; its scores are taken from the WOD-E2E leaderboard. Qwen-Drive-1.0~\citep{zhou2026qwen} couples Qwen3.5-4B with a flow-matching planner conditioned on VLM representations of the scene and an autoregressively generated text reasoning trace; its RL stage, like ours, optimizes only the planner with PDMS and RFS rewards. Fast-dDrive~\citep{zhang2026fast} is a block-diffusion dLLM that generates reasoning and trajectory as structured text and reports Waymo results only. Both reason in text where VFlash uses a visual plan. Latency measurement conditions for these methods, including our own measurement of Qwen-Drive-1.0, are given in Appendix~\ref{app:latency_eval}.

\end{document}